\documentclass[10pt,twocolumn,letterpaper]{article}

\usepackage{cvpr}              
\usepackage[accsupp]{axessibility}
\definecolor{cvprblue}{rgb}{0.21,0.49,0.74}
\usepackage[pagebackref,breaklinks,colorlinks,allcolors=cvprblue]{hyperref}
\usepackage{amsmath}
\usepackage{amsfonts}
\usepackage{bbm}
\usepackage{multirow}
\usepackage{booktabs}
\usepackage{float}
\usepackage{colortbl}
\usepackage{stfloats} 

\usepackage{pifont}
\newcommand{\cmark}{\ding{51}}%
\newcommand{\xmark}{\ding{55}}%

\title{Dual Diffusion for Unified Image Generation and Understanding}
\author{
Zijie Li$^{1}$\thanks{The first two authors contributed equally to this work, work done during an internship at ByteDance.} \quad Henry Li$^{2}$$^*$ \quad Yichun Shi$^{3}$ \quad Amir Barati Farimani$^{1}$ \quad Yuval Kluger$^{2}$ \\ Linjie Yang$^{3}$ \quad Peng Wang$^{3}$  \\
zjli.jlee@gmail.com, henry.li@yale.edu \\ \{linjie.yang, peng.wang\}@bytedance.com
\\
\small{$^{1}$Carnegie Mellon University~~~~~~$^2$Yale University~~~~~~$^3$ByteDance Seed} \\
\href{https://zijieli-jlee.github.io/dualdiff.github.io/}{Project website}
}

\begin{document}
\maketitle
\begin{abstract}
Diffusion models have gained tremendous success in text-to-image generation, yet still struggle with visual understanding tasks, an area dominated by autoregressive vision-language models.
We propose a large-scale and fully end-to-end diffusion model for multi-modal understanding and generation that significantly improves on existing diffusion-based multimodal models, and is the first of its kind to support the full suite of vision-language modeling capabilities.
Inspired by the multimodal diffusion transformer (MM-DiT) and recent advances in discrete diffusion language modeling, we leverage a 
cross-modal maximum likelihood estimation framework that simultaneously trains the conditional likelihoods of both images and text jointly under a single loss function, which is back-propagated through both branches of the diffusion transformer. The resulting model is highly flexible and capable of a wide range of tasks including image generation, captioning, and visual question answering. Our model attained competitive performance compared to recent unified image understanding and generation models, demonstrating the potential of multimodal diffusion modeling as a promising alternative to autoregressive next-token prediction models. 
\end{abstract}
\section{Introduction}
\label{sec:intro}


We are currently in the midst of a multimodal generative modeling revolution. Large scale diffusion models such as Stable Diffusion \citep{esser2024scaling}, Dall-E \citep{ramesh2022dalle2}, FLUX, and Imagen~\citep{saharia2022imagegen} have become indisputable industry leaders for generating high fidelity images from text descriptions, enabling the accurate modeling and sampling of complex and high dimensional distributions of images given text. Conversely, autoregressive next-token prediction models have achieved groundbreaking performance both in pure text generation and reasoning such as in ChatGPT \citep{achiam2023gpt}, Gemini \citep{team2023gemini}, and Llama \citep{dubey2024llama} and in visually-grounded text generation with large language models (LLMs), as seen with LLaVA \citep{liu2024visual} or BLIP-2 \citep{li2023blip}.

Given these developments, a natural question comes to mind: \textit{Can these existing image-to-text (I2T) or text-to-image (T2I) systems be modified to reason with and generate data in the \textit{reverse} direction?} A positive answer would suggest the possibility of producing a fully multimodal model that is able to understand and sample from conditional distributions between modalities in an omni-directional manner. Moreover, unifying these generative frameworks under a single model with shared parameters can confer a multitude of downstream benefits including improved reasoning, simplified implementation, and may be a natural next step towards artificial general intelligence \citep{huh2024platonic,tong2024cambrian}.

With autoregressive next-token prediction models, this query has already been answered resoundingly in the affirmative, as evidenced by a multitude of studies \citep{team2024chameleon,gao2024lumina,yu2023cm3leon,sun2023emu,wang2024emu3, ge2023seed,dong2024dreamllm} demonstrating T2I capabilities of finetuned LLMs. This is in part due to the known next-token generative capability of autoregressive models with visual tokens \citep{yu2023language,tian2024visual,li2024autoregressive}. 

On the contrary, with diffusion models there has been surprisingly little evidence of a similar reverse capacity. Until recently, generative diffusion models have struggled with language modeling due to the lack of an empirically performant discrete diffusion process on text tokens, in spite of continued research in this area \citep{austin2021structured,dieleman2022continuous,li2022diffusionlm}. 
At present, multimodal diffusion models either exhibit limited text reasoning capabilities and partial text diffusion \citep{bao2023unidiffuser,xu2023versatile}, which require an autoregressive model such as GPT2 \citep{radford2019language} to decode denoised text latents, or emerge as add-ons to pretrained LLMs fine-tuned in conjunction with a diffusion loss \citep{zhou2024transfusion,xie2024show}, and ultimately still rely entirely on next-token prediction for text generation.


We leverage the novel progress in this domain to revisit the above-mentioned question and propose a dual-branch diffusion model based on the multimodal diffusion transformer (MM-DiT) architecture \citep{esser2024scaling}, which we modify to output diffusion targets on both modalities of the neural network. We then train our model to perform continuous latent space diffusion on the image branch and discrete masked token diffusion on the text branch. Our novel implementation also allows for controllable infilling in the token space, enabling visual question answering and vision language assistance, which prior diffusion-based models were incapable of. To the best of our knowledge, this is the first end-to-end multimodal diffusion model fully capable of full-featured I2T and T2I generation. 

Moreover, we demonstrate the compatibility of our framework with existing diffusion foundation models such as Stable Diffusion 3 (SD3) \citep{esser2024scaling}, allowing us to initialize our model with pretrained checkpoints, and reveals remarkably fast adaptation capabilities of the proposed architecture on text generation, producing meaningful text output in under 25B text tokens when initialized with an SD3 checkpoint. 
Our contributions can be summarized as follows:
\begin{itemize}
    \item We introduce a fully end-to-end cross-modal diffusion model that unifies image and text diffusion under a single transformer, which to the best of our knowledge is the first of its kind.
    \item We propose a simple, elegant, and easy to implement joint loss function that simultaneously trains the conditional text and image modalities in a unified, end-to-end fashion.
    \item We demonstrate performance on an expanded set of multimodal tasks including image generation, visual captioning, and visual question answering using a diffusion-only model, significantly improving on the capabilities and performance of prior multimodal diffusion models.
\end{itemize}

\begin{table}[h]
\centering
\resizebox{\linewidth}{!}{
\begin{tabular}{lccccc}
        \toprule
	& \multicolumn{2}{c}{Modality} & \multicolumn{3}{c}{Task} \\
        \cmidrule(r){2-3}\cmidrule(lr){4-6}
             & Image & Text & Image & Image & Visual Question \\
             & Backbone & Backbone & Gen & Cap. & Answering \\
            \hline
Versatile Diffusion \citep{xu2023versatile} & \textbf{Diffusion} & Diff. + AR & \cmark & \cmark & \xmark \\
            Unidiffuser \citep{bao2023unidiffuser} & \textbf{Diffusion} & Diff. + AR & \cmark & \cmark & \xmark \\
            Show-O \citep{xie2024show} & \textbf{Diffusion} & AR & \cmark & \cmark & \cmark \\
            Transfusion \citep{zhou2024transfusion} & \textbf{Diffusion} & AR & \cmark & \cmark & \cmark \\
	    Ours & \textbf{Diffusion} & \textbf{Diffusion} & \cmark & \cmark & \cmark \\
            \bottomrule
\end{tabular}
}
\caption{A side-by-side comparison between the backbones and supported features of our work compared to those of existing diffusion-based multimodal methods.}
\vspace{-2mm}

\end{table}




\section{Background}
\label{sec:background}

In this section, we review the basic concepts that underpin our proposed model. Generally, diffusion models \citep{ho2020denoising,song2020score} are inspired by non-equilibrium thermodynamics \citep{sohl2015deep} designed to evaluate a likelihood $p_\theta(\mathbf{x}) = \int p_\theta(\mathbf{x}_{0:T}) d\mathbf{x}_{1:T}$ where data $\mathbf{x}_0 := \mathbf{x}$ are related to a set of latent variables $\mathbf{x}_{1:T}$ by a diffusion process that gradually corrupts the original data.



\subsection{Continuous Diffusion}
\label{sec:continuous_diffusion}
Continuous diffusion models operate on continuous vectors by learning to reverse the noise-corruption forward process
\begin{equation}
    \label{eq:continuous diffusion xt}
    \mathbf{x}_t = \alpha_t \mathbf{x} + \sigma_t \boldsymbol{\epsilon},
\end{equation}
parameterized by time-dependent scalar $\alpha_t$ and $\sigma_t$, where $\alpha_t, \sigma_t>0$, $\alpha_t/\sigma_t$ decreases monotonically, and $\boldsymbol{\epsilon}$ is an appropriately selected \textit{i.i.d.} noise variable. In score-based diffusion models \citep{ho2020denoising, song2020score}, $\alpha_t, \sigma_t$ are determined by a forward stochastic differential equation (SDE) that pushes $\mathbf{x}_t$ towards $\mathcal{N}(0, I)$ as $t \mapsto \infty$. New samples can be generated by learning the reverse process through estimating the score function \citep{anderson1982reverse, vincent2011connection, song2020score} $\nabla\log p_t(\mathbf{x}_t)$. Alternatively, from \eqref{eq:continuous diffusion xt}, the following ordinary differential equation (ODE) can be derived:

\begin{equation}
    \label{eq:flow velocity}
    \dot{\mathbf{x}}_t = \mathbf{v}(\mathbf{x}_t, t),
\end{equation}
with velocity field $\mathbf{v}(\mathbf{x}_t, t)=\dot{\alpha_t}\mathbf{x}+\dot{\sigma_t}\mathbf{\epsilon}$. The ODE in \eqref{eq:flow velocity} pushes the distribution of $\mathbf{x}_t$ from $p_0$ to $p_T$. To generate new samples, we can use neural networks to approximate $\mathbf{v}$ and then integrate ODE \eqref{eq:flow velocity} backward in time starting from $\mathbf{x}_T \sim \mathcal{N}(\mathbf{0}, \mathbf{I})$. A common choice of $\alpha_t, \sigma_t$ in flow matching model is $\alpha_t=1-t, \sigma=t$ and therefore $\mathbf{v}=\epsilon-\mathbf{x}$, which corresponds to the optimal transport interpolant between two distribution $p_0$ and $p_1$ \citep{lipman2022flow, liu2023flow}. The neural network for regressing the velocity field $\mathbf{v}$ in \eqref{eq:flow velocity} is trained by optimizing the flow matching loss
\begin{equation}
\label{eq:fm loss}
L_{\text{FM}}=\mathbb{E}_{t,q(\mathbf{x}_t|\mathbf{x})}||\mathbf{v}_\theta(\mathbf{x}_t, t)-(\epsilon-\mathbf{x})||_2^2.
\end{equation}

Recent work such as Stable Diffusion 3 \citep{esser2024scaling} has demonstrated the superiority of flow matching model on text-to-image generation, thus in this work we adopt flow matching objective for modeling the distribution of images.

\subsection{Discrete Diffusion}
\label{sec:discrete diffusion}
In discrete diffusion, the variate $\mathbf{x} \in \mathcal{X} \times \dots \times \mathcal{X}$ has finite support over the product space of $\mathcal{X} = \{1, \dots, N\}$, where in language models $N$ is the vocabulary size of the token embedding. Generally, there are two ways to approach this modeling task. The first line of works \citep{li2022diffusion,chen2022analog,dieleman2022continuous,lovelace2024latent,gulrajani2024likelihood} apply a continuous relaxation to the discrete variable and proceed with a continuous reformulation of the framework, allowing the application of the equations in Section \ref{sec:continuous_diffusion}. This greatly simplifies the diffusion modeling itself, but introduces a significant source of error in the mapping between discrete and relaxed continuous states. Conversely, the diffusion process is extended to the discrete token space \citep{austin2021structured,meng2022concrete,lou2023discrete,sahoo2024simple}, which removes the need for the aforementioned mapping via a specialized discrete diffusion formulation. In our work, we will focus on this latter perspective given its empirical potential as validated in recent works \citep{lou2023discrete,sahoo2024simple,gat2024discrete,shi2024maskdiff}. Leveraging continuous-time Markov chain (CTMC) theory, the marginal distributions $p_t$ can be described by a family of linear ordinary differential equations
\begin{equation}
    \frac{d p_t}{dt} = Q_t p_t,
    \label{eq:discretized_fokker_planck}
\end{equation}
where $p_0 \approx p_\text{data}$ and $p_1 = p_\text{stationary}$, and $Q_t$ is a time-dependent sequence of transition matrices that provides a mapping between the two distributions. We consider the case of absorbing state (i.e., masked) diffusion that are shown to work well on text modeling \citep{lou2023discrete, sahoo2024simple, shi2024maskdiff}. This formulation induces the posterior ($0<s<t$)
\begin{equation}
    q(\mathbf{x}_s | \mathbf{x}_t, \mathbf{x}) = \begin{cases}
        \text{Cat}(\mathbf{x}_s | \mathbf{x}_t) & \mathbf{x}_t \neq \mathbf{m} \\
        \text{Cat}(\mathbf{x}_s | \frac{(1 - \alpha_s) \mathbf{m} + (\alpha_s - \alpha_t)\mathbf{x}}{1 - \alpha_t}) & \text{o.w.} \\
    \end{cases}
    \label{eq:mdlm_posterior}
\end{equation}
where clean data $\mathbf{x}$ is a discrete variable (one-hot vector) with $N$ categories,
with the marginal
\begin{equation}
    \label{eq:mdlm xt}
    q(\mathbf{x}_t | \mathbf{x}) = \text{Cat}[\mathbf{x}_t | \alpha_t \mathbf{x} + (1 - \alpha_t) \mathbf{m}],
\end{equation}
where $\text{Cat}(\cdot|\boldsymbol{\pi})$ denotes the categorical distribution over different classes with probability $\boldsymbol{\pi}$, and $\mathbf{m}$ denotes the mask absorbing state.

To reverse this process, one may either model the density ratio $s_\theta(\mathbf{x})_\mathbf{y} \approx \frac{p_t(\mathbf{x})}{p_t(\mathbf{y})}$ given two sequences $\mathbf{x}, \mathbf{y} \in \mathcal{X} \times \dots \times \mathcal{X}$ as in \citep{lou2023discrete}, or the denoised variate $\mathbf{x}_\theta(\mathbf{x}_t, \alpha_t) \approx \mathbf{x}$ directly as in \citet{sahoo2024simple, shi2024maskdiff}. In the former, the modeled density ratios induce a specialized reverse transition matrix $\bar{Q}_t$ that can be leveraged in Eq. \ref{eq:discretized_fokker_planck}. In the latter, $\mathbf{x}_\theta$ can be directly substituted for $\mathbf{x}$ in Eq. \ref{eq:mdlm_posterior}.
In this work, we follow \citet{sahoo2024simple} that enforces zero-probability on the mask state $\mathbf{m}$ and keeps all un-masked state unchanged during reverse sampling. This induce a simplified (negative) variational lower bound under the continuous time limit
\begin{equation}
\label{eq:mdlm elbo}
    L_{\text{NELBO}} = \mathbb{E}_{q(\mathbf{x}_t|\mathbf{x})}
    \left[
    \int_{0}^1 \frac{\alpha_t'}{1-\alpha_t}\log(\mathbf{x}_\theta(\mathbf{x}_t, \alpha_t) \cdot \mathbf{x}) dt
    \right].
\end{equation}
In practice, we can use Monte-Carlo sampling to approximate and evaluate the loss function in \eqref{eq:mdlm elbo}. Following \citet{sahoo2024simple}, we use a log-linear schedule: $\alpha_t = 1-t$.

\begin{figure*}[t]
    \centering
    \includegraphics[width=\linewidth]{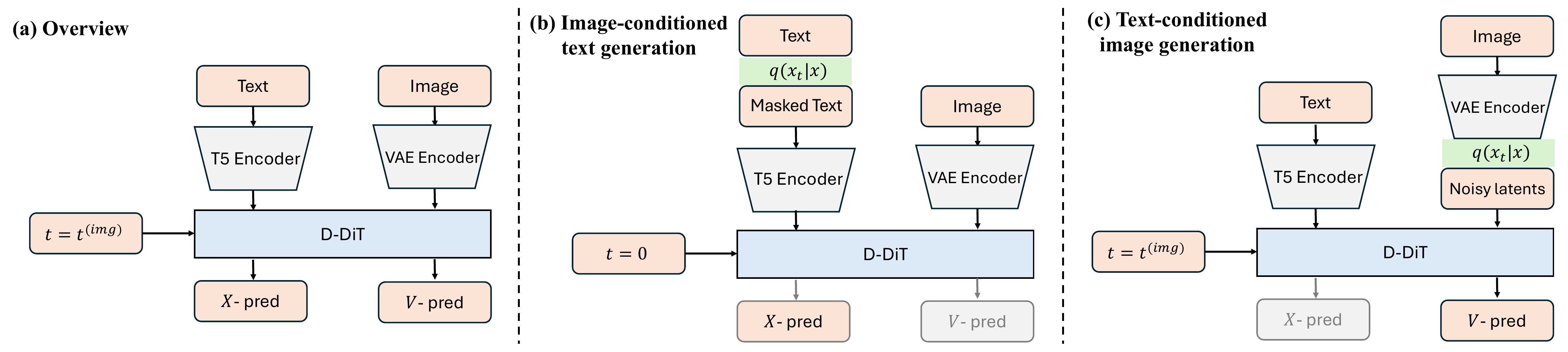}
    \caption{Our proposed model, the Dual Diffusion Transformer (D-DiT) that simultaneously models image and text distributions via a joint denoising diffusion training loss. \textbf{a) }An overview of the model architecture. The gray blocks (T5 encoder, image autoencoder) are kept fixed throughout training and inference. \textbf{b)} During training for (image-conditioned) text denoising, the text input is randomly masked while the image is noise-free. \textbf{c)} During training for text-conditioned image denoising, the image is randomly noised while the text is noise-free.}
    \label{fig:overview d-dit}
    \vspace{-3mm}
\end{figure*}

\section{Method}

\label{sec:methods}

We propose an end-to-end multi-modal diffusion model named Dual Diffusion Transformer (D-DiT) with a unified backbone that jointly models image and text distribution. More specifically, given image $\mathbf{x}^{(\text{img})}$ and text $\mathbf{x}^{(\text{txt})}$, we are interested in modeling the conditional distribution $p(\mathbf{x}^{(\text{img})}|\mathbf{x}^{(\text{txt})})$ and $p(\mathbf{x}^{(\text{txt})}|\mathbf{x}^{(\text{img})})$. The former is usually referred to as text-to-image generation and the latter forms the basis for various image understanding tasks such as captioning and visual question answering. 
\vspace{-2mm}

\subsection{Architecture}
 Inspired by the MM-DiT in SD3~\citep{esser2024scaling}, our proposed D-DiT is a Transformer-based model comprising two branches - one for processing image tokens and another for processing text tokens. The image and text tokens attend to each other in every attention layer. In D-DiT, the output of the image branch is the prediction of velocity defined in \eqref{eq:flow velocity} with text conditioning, while the output for the text branch is the $\mathbf{x}^{(\text{txt})}$ prediction with image conditioning. The scalar timestep embedding modulates every layer's feature map via AdaLN (adaptive layernorm) \citep{peebles2023dit}. We only input the timestep information $t$ to the model for image generation, as $\mathbf{x}^{(\text{txt})}_t$ implicitly encodes this information as the ratio of masked tokens in the sequence. In addition, we add a text encoder with bi-directional attention on top of the text branch of the diffusion model. While the asymmetry between image and text branches is not strictly required, having a text encoder on top of a DiT model allows us to easily adapt many existing text-to-to-image models such as SD3 and FLUX as pretrained backbones for our D-DiT model (see Table 9 in the Appendix for a comparison). Note that the text encoder should not use a causal mask as this will violate the masked diffusion process. 

 To reduce the computational cost associated with modeling high-resolution images, we follow prior works on latent-space (image) diffusion \citep{rombach2022ldm}, which encode images from the raw pixel space into a spatially compressed latent space obtained from a variational autoencoder (VAE) trained with a discriminator loss \citep{esser2021taming} and KL-divergence regularization \citep{kingma2013auto}.

\subsection{Training}
 We propose a combined training objective for image-text modeling, which is essentially a joint denoising target that combines continuous and discrete diffusion. Formally, we use flow matching introduced in Section \ref{sec:continuous_diffusion} to learn the conditional distribution of images and masked diffusion introduced in Section \ref{sec:discrete diffusion} to learn the conditional distribution of texts. During training, corrupted samples $\mathbf{x}_{t^{(\text{img})}}, \mathbf{x}_{t^{(\text{txt})}}$ \footnote{For better readability superscript $(\text{txt}),(\text{img})$ are omitted.} are drawn from the corresponding forward corruption processes $q(\mathbf{x}_t|\mathbf{x})$ defined in \eqref{eq:continuous diffusion xt} and \eqref{eq:mdlm xt} respectively. We then calculate the diffusion loss for each modality as
 \vspace{-2mm}
 \begin{equation}
 \label{eq:joint diffusion loss}
    \begin{aligned}
     &L_{\text{image}} \\
     &= \mathbb{E}_{t,q^{(\text{img})}}
     \left|\left|
     \mathbf{v}_{\theta}\left(
     \mathbf{x}^{(\text{img})}_{t}, t, \mathbf{x}^{(\text{txt})}
     \right)
     - (\boldsymbol{\epsilon} - \mathbf{x}^{(\text{img})})
     \right|\right|_2^2, \\
     &L_{\text{text}} \\
     &= \mathbb{E}_{q^{(\text{txt})}}
     \left[
     -\frac{1}{K}\sum \nolimits_{i=1}^{K}
     \log [
     \mathbf{x}_{\theta}(\mathbf{x}_{t_i}^{(\text{txt} )}, \mathbf{x}^{(\text{img})})
     \cdot\mathbf{x}
     ] / t_i\right],
 \end{aligned}  
 \end{equation}
 In text diffusion, we use antithetic sampling \citep{kingma2021variational} for timesteps $t_i$ by discretizing $(\delta, 1]$ into $K$ points uniformly with $\delta$ being a small number to avoid numerical instability. In image diffusion, we sample $t$ from the log-normal distribution. We do not corrupt the conditioning samples during training, i.e., the image diffusion timestep is always set to zero when predicting text distribution and vice versa.

 In summary, the overall dual modality training loss is a simple weighted combination of the above single modality diffusion loss:
 \begin{equation}
     L_{\text{dual}} = L_{\text{image}} + \lambda_{\text{text}} L_{\text{text}},
 \end{equation}
with $\lambda_{\text{text}}$ being a hyperparameter.

\subsection{Inference}
We introduce three types of sampling-based inference which can be used for different vision-language tasks, which we detail below. 
\vspace{-2mm}

\paragraph{Text-to-image Generation} To perform text-guided image generation, i.e. $\mathbf{x} \sim p(\mathbf{x}^{(\text{img})} | \mathbf{x}^{(\text{txt})})$, we use the commonly adopted classifier-free guidance (CFG) technique \citep{ho2022classifier} to sample from the conditional distribution $p(\mathbf{x}^{(\text{img})}_{t} | \mathbf{x}^{(\text{txt})})$, which amounts to a re-weighting of the velocity prediction
\begin{equation}
    \tilde{v}_t = s \mathbf{v}_{\theta}\left(
     \mathbf{x}^{(\text{img})}_{t}, t, \mathbf{x}^{(\text{txt})}
     \right) + (1-s)\mathbf{v}_{\theta}\left(
     \mathbf{x}^{(\text{img})}_{t}, t, \emptyset
     \right),
\end{equation}
where $s$ is a hyperparameter that controls the scale of guidance and $\emptyset$ is a suitable null embedding (e.g. the embedding of an empty text) .
\vspace{-2mm}

\paragraph{Image-to-text Generation} To sample images from the conditional distribution, we can use ancestral sampling to draw from the posterior distribution $q(\mathbf{x}_s| \mathbf{x}_t, \mathbf{x})$ in \eqref{eq:mdlm_posterior} by plugging in prediction $\mathbf{x}\approx\mathbf{x}_\theta(\mathbf{x}_t^{(\text{txt})}, \mathbf{x}^{(\text{img})};t=0)$.
\vspace{-2mm}

\paragraph{Image-to-text In-filling} In certain tasks, both text conditioning information and image conditioning information are available, such as in a visual question answering task where an image and an associated question are provided. For such cases, we would like to sample $\mathbf{x} \sim p(\mathbf{x}^{(\text{answer})} | \mathbf{x}^{(\text{img})}, \mathbf{x}^{(\text{question})})$.

\begin{figure}[h]
    \centering
    \includegraphics[width=\linewidth]{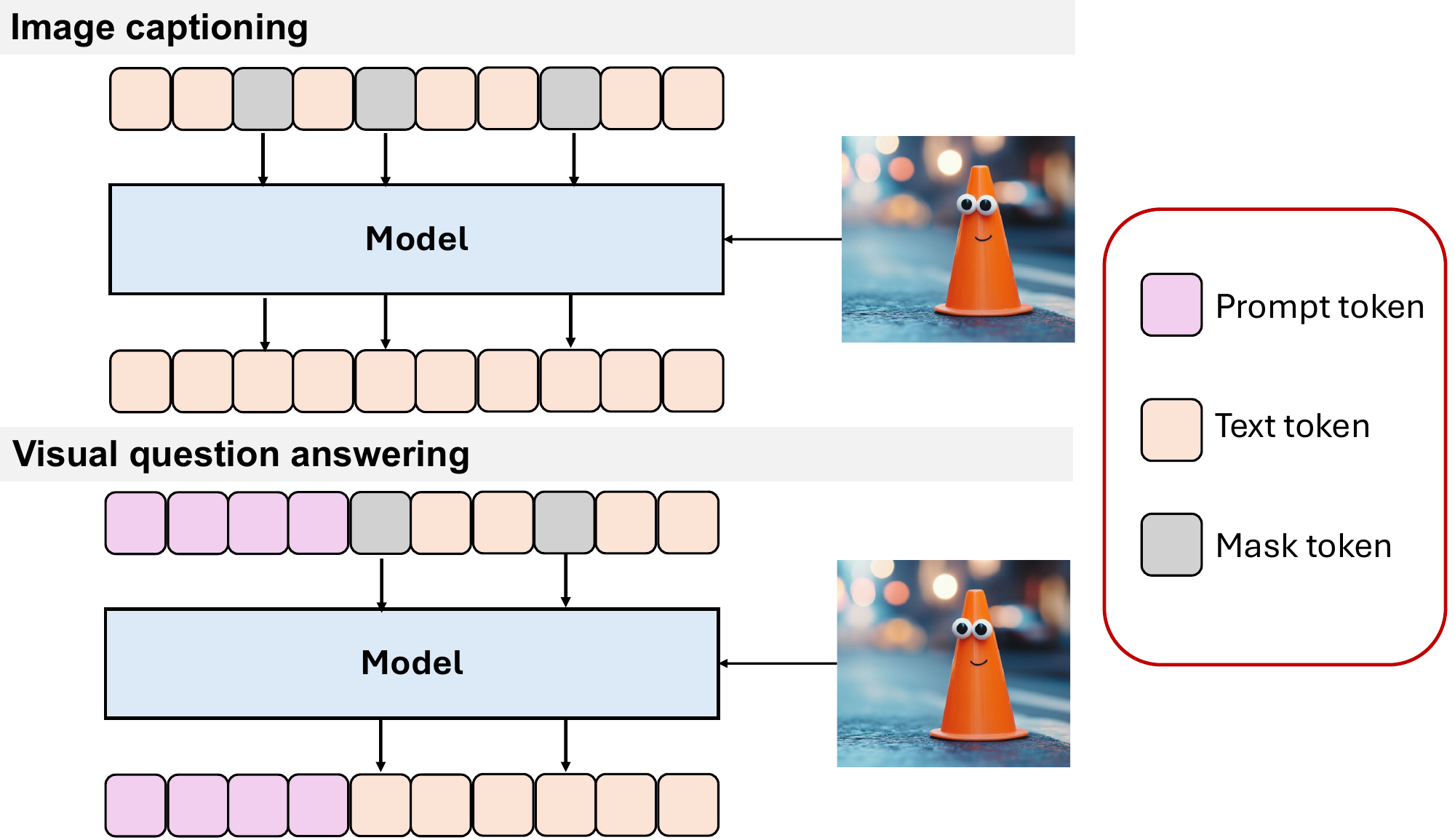}
    \vspace{-1mm}
    \caption{Text masking during both training and sampling under the image captioning (above) and visual question answering (below) tasks with our proposed model.}
    \label{fig:text sampling flowchart}
    \vspace{-2mm}
\end{figure}

To perform this task, we initialize the diffusion prior of the question with masked tokens and leverage the robust text in-filling capabilities of the text diffusion model to complete the sequence by sampling from the conditional distribution. The text question tokens are kept fixed throughout sampling (Figure \ref{fig:text sampling flowchart}).

\section{Experiments}
\label{sec:experiments}

\begin{figure*}[t]
    \centering
    \includegraphics[width=\linewidth]{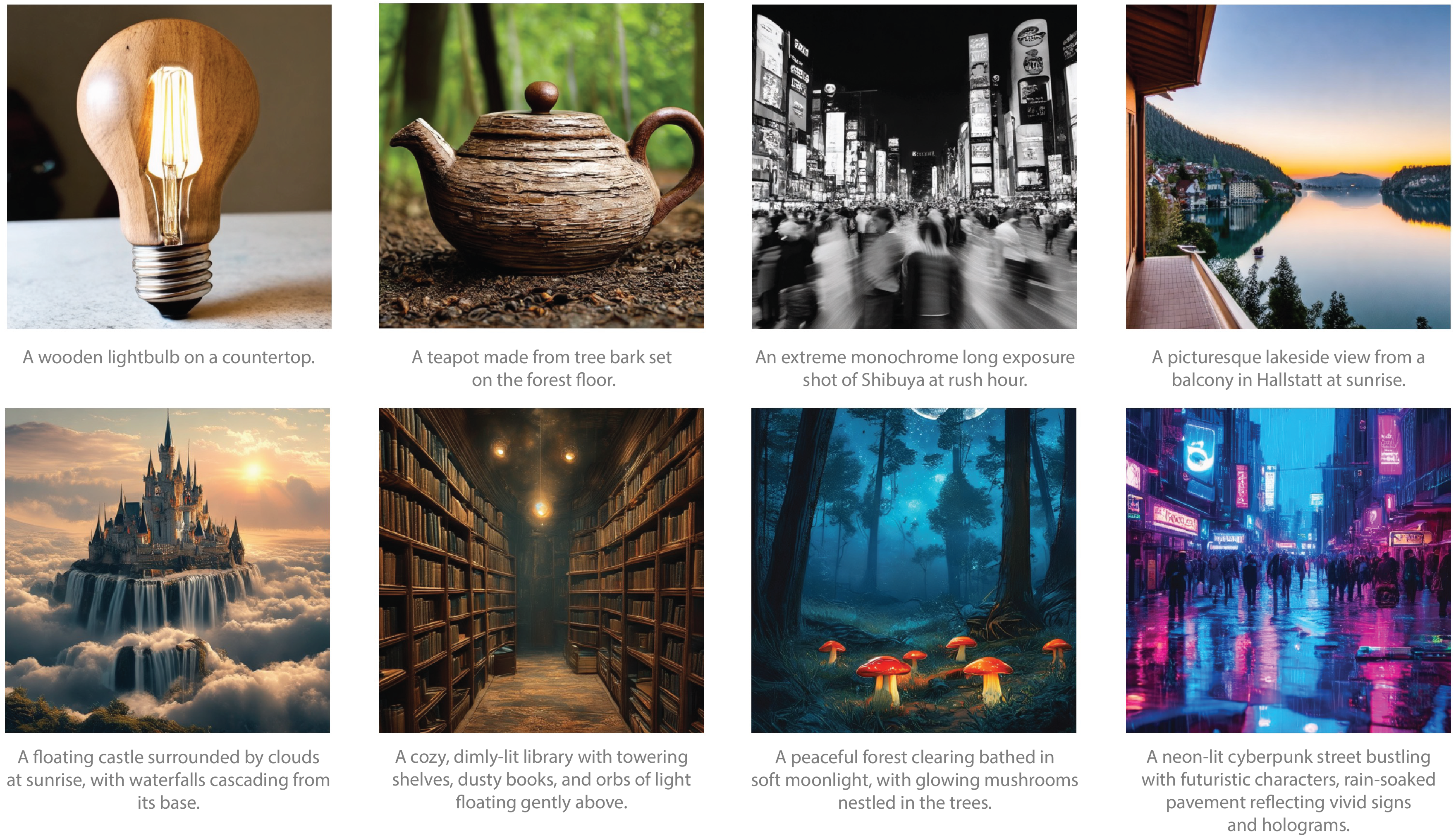}
        \vspace{-7mm}
    \caption{Text-to-image samples generated from the model. We draw images from the reverse diffusion process via the Euler solver with $T=28$ diffusion steps.}
    \label{fig:main_t2i}
    \vspace{-3mm}
\end{figure*}

\subsection{Experimental Setup}
\paragraph{Implementation details}
We implement our proposed framework based on the open-sourced SD3-medium model~\citep{esser2024scaling}\footnote{https://huggingface.co/stabilityai/stable-diffusion-3-medium}. We initialize the model weights of the DiT from the pretrained checkpoint and add a linear head on top of the text branch for text denoising. Following SD3, we adopt the existing T5 encoder/tokenizer \citep{raffel2023t5}, and SD3's image VAE, whose weights remain unchanged throughout all the experiments (except for the mask token embedding in T5). We remove the CLIP text encoders in the SD3 model due to its causal attention mask and for a simplified model structure. We use the special token \verb+<extra_id0>+ in T5's vocabulary to represent the mask token in masked diffusion, as this token is used to mark the masked token in the mask pretraining process of original T5 model. This way, we find the model can generate text reasonably well even without updating the weight of this token embedding. To further reduce the domain gap, we unfreeze the token embedding of \verb+<extra_id0>+ during the second stage of the training. 

Different from multi-modal models that are built upon language models, our model has never been trained on text-only generation. In preliminary experiments, we found that adding a text-only target (i.e. unconditional text generation) to the model does not influence its captioning performance significantly. An interesting future direction can be extending the proposed framework to model the marginal distribution of each modality.
\vspace{-3mm}
\paragraph{Datasets}
We train the model in three stages on publicly available datasets. The total number of image-text pairs used is roughly 40M. We list the details of the dataset and training setup for each stage below, where all the training stages use the joint diffusion loss defined in \eqref{eq:joint diffusion loss}.
\begin{enumerate}
    \item \textbf{Dual diffusion pretraining}. The original SD3 model was only trained on ambient image-text pairs, and not solely on text data itself. To adapt D-DiT to text generation tasks, we train it on the joint diffusion loss for 60K iterations with a batch size of 512. The maximum text token length is truncated to 64 and we use an image resolution of 256. The dataset used in this stage is re-captioned Datacomp-1b \citep{gadre2023datacomp, li2024recapdatacomp} (the model has only seen around 30M images in this stage, which is less than 
    $3\%$ of the total images in the dataset). 
    \item \textbf{Continued pretraining on higher quality data}. 
    We then unfreeze the masked token embedding in T5 and train the model for 200k iterations on an image understanding dataset with rich textual description, which consists of the pretraining dataset from ShareGPT4V\citep{chen2023sharegpt4v} (1.3M images) and OpenImages (1.9M subset with object detection annotations) \citep{kuznetsova2020openimages} re-captioned by ShareCaptioner\footnote{https://huggingface.co/Lin-Chen/ShareCaptioner}. The text token length is set to 256 and image resolution to 256, with a batch size of 512. Finetuning the mask token embedding reduces the domain gap as T5 encoder has not seen sequences filled with a high percentage of mask tokens during its pretraining.
    
    However, as updating the mask token embedding requires backpropagating through the large T5 encoder, we freeze the mask embedding after this round of training on the image understanding dataset. We observe that the $\ell2$ difference between the mask token embedding from different training iterations does not change much after 100k iterations. 
    
    Here, we may conduct an optional high resolution model finetuning on the aforementioned image understanding dataset together with a higher quality dataset with 10M images (9M re-captioned LAION-1024 and 1M midjourney images\footnote{https://huggingface.co/datasets/CaptionEmporium/midjourney-niji-1m-llavanext}). In this training stage, the image diffusion loss is calculated on the high quality image dataset whereas the text diffusion loss is calculated on the understanding dataset. We finetune the model for 80k iterations, with image resolution 512, text token length 256, and a batch size of 768. Only our 512$\times$512 model variant requires this training stage.
    \item \textbf{Visual instruction tuning}. Finally, we finetune our model on a medley of instruction-tuning datasets to promote joint text-image conditioned text generation. We combine the LLaVA-Pretrain558K and LLaVA-v1.5-mix-665K visual instruction tuning datasets with the training splits for TextVQA and VizWiz and train for 50k iterations. 
    Following the convention in LLaVA-1.5, the model is trained to distinguish between long-form and short answers, multiple choice answers, or captions via task-specific instruction prompts that come after the question, e.g. \textit{"Answer the question using a single word or phrase,"} or \textit{"Describe the image concisely."}
\end{enumerate}




\subsection{Multi-modal Understanding}
Existing multi-modal diffusion models such as UniDiffuser \citep{bao2023unidiffuser} and Versatile Diffusion \citep{xu2023versatile} performed text diffusion in a CLIP latent space, which hampered their ability to perform text completion, a necessary feature for general question answering and conversation-based tasks. This is no longer a limitation with our proposed D-DiT due to its discrete masked diffusion branch, allowing us to leave question tokens unmasked throughout sampling. We are thus able to evaluate our fine-tuned model on a full suite of image-to-text generation tasks, including image captioning and visual question answering benchmarks, as well as long-form visual assistance responses.

We first evaluate the visual understanding capabilities of D-DiT via the academic question answering benchmarks VQAv2 \citep{balanced_vqa_v2}, VizWiz \citep{bigham2010vizwiz}, OKVQA \citep{marino2019ok}, GQA \citep{hudson2019gqa}, POPE \citep{Li-hallucination-2023}, as well as MME \citep{fu2023mme}. Due to the short-form nature of the questions, we perform sampling with 16 diffusion steps, and compare against a selection of multi-modal models, including I2T only and I2T + T2I models. Our results are summarized in Table \ref{table:lm_benchmarks}. We note that our D-DiT is the only diffusion-only multi-modal model capable of visual question answering tasks, already boosting performance that is competitive with recent I2T + T2I models. Our model at 512 resolution outperforms Show-O on MME, GQA, and POPE, approaching the performance of auto-regressive VLMs such as QWEN-VL and BLIP-2.

\begin{figure*}[t]
    \centering
    \includegraphics[width=\linewidth]{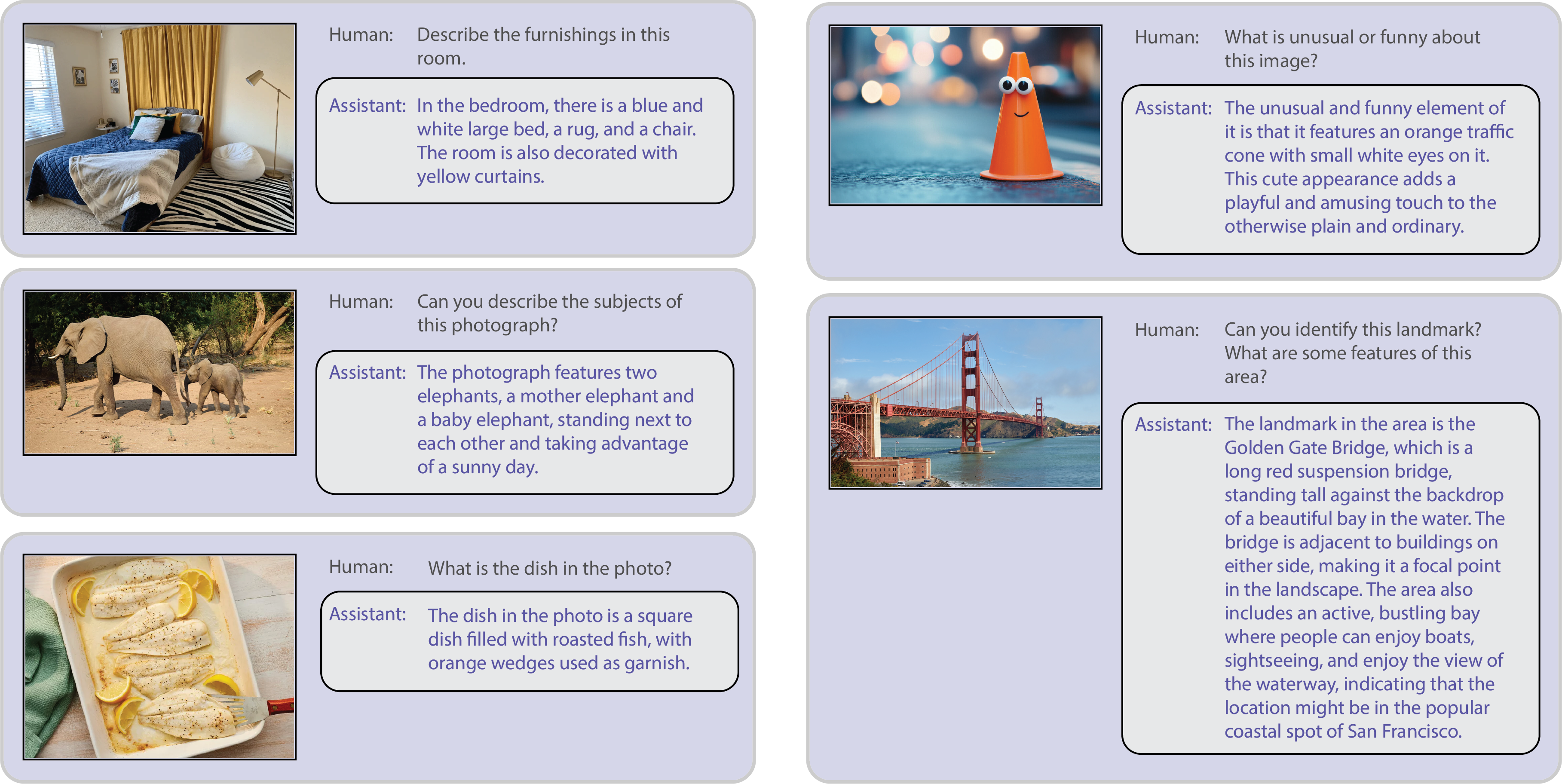}
        \vspace{-5mm}
    \caption{Multi-modal dialogue examples generated from our model. To our knowledge, D-DiT is the first diffusion-based multimodal model capable of instruction-based vision and language conversation.}
    \label{fig:main_i2t}
\end{figure*}

Next, we provide qualitative examples of the D-DiT, providing images and gauging the model's visual language assistance capabilities via image-related queries. Given the longer format of the responses, we sample D-DiT responses with 256 diffusion steps. Our model provides answers to human queries in a manner that suggests a fine-grained multi-modal understanding of the image and text conditioning (Figure \ref{fig:main_i2t}). 

\subsection{Text-to-image Generation}

Besides the image-conditioned text generation, we also test model's text-to-image generation capability. Following previous works, we evaluate our 512$\times$512 model after the second training stage on the GenEval benchmark, which measures model's prompt following capability \citep{ghosh2023geneval}. We follow the default setting in the open-sourced SD3 checkpoint where we use a Euler solver with 28 sampling steps and a CFG scale of $7.0$. We observe that the joint diffusion training does not cause catastrophic forgetting on the model; the fine-tuned D-DiT preserves the performance of the original SD3 model and slightly improves on some metrics such as color accuracy after joint training. Qualitative evaluation samples are shown in Figure \ref{fig:main_t2i}, where we observe that the ability to generate highly aesthetic images is preserved.

\begin{table}[h]
\centering
\setlength{\tabcolsep}{0.75pt}
\scalebox{0.75}{
\begin{tabular}{ccccccccc} 
\cmidrule[\heavyrulewidth]{1-3}\cmidrule[\heavyrulewidth]{4-9}
\multirow{2}{*}{Model} & \multirow{2}{*}{\begin{tabular}[c]{@{}c@{}}params \\(B)\end{tabular}} & \multirow{2}{*}{Overall} & \multicolumn{2}{c}{Objects} & \multirow{2}{*}{Counting} & \multirow{2}{*}{Colors} & \multirow{2}{*}{Position} & \multirow{2}{*}{\begin{tabular}[c]{@{}c@{}}Color \\attribution\end{tabular}} \\ 
\cmidrule{4-5}
 &  &  & \small Single & \small Two &  &  &  &  \\ 
\midrule
PixArt-$\alpha$ \citep{chen2023pixart} & 0.6 & 0.48 & 0.98 & 0.50 & 0.44 & 0.80 & 0.08 & 0.07 \\
SD V2.1 & 0.9 & 0.50 & 0.98 & 0.51 & 0.44 & 0.85 & 0.07 & 0.17 \\
DALL-E 2 \citep{ramesh2022dalle2} & 6.5 & 0.52 & 0.94 & 0.66 & 0.49 & 0.77 & 0.10 & 0.19 \\
SDXL \citep{podell2023sdxl} & 0.9 & 0.55 & 0.98 & 0.74 & 0.39 & 0.85 & 0.15 & 0.23 \\
DALL-E 3 & - & 0.67 & 0.96 & 0.87 & 0.47 & 0.83 & 0.43 & 0.45 \\ 
\midrule
CoDI \citep{tang2024codi}& - & 0.31 & 0.89 & 0.16 & 0.16 & 0.65 & 0.02 & 0.01 \\
LWM \citep{liu2024lwm} & 7 & 0.47 & 0.93 & 0.41 & 0.46 & 0.79 & 0.09 & 0.15 \\
SEED-X \citep{ge2023seed} & 17 & 0.49 & 0.97 & 0.58 & 0.26 & 0.80 & 0.19 & 0.14 \\
Chameleon \citep{team2024chameleon} & 7 & 0.39 & - & - & - & - & - & - \\
Show-O \citep{xie2024show} & 1.3 & 0.68 & 0.98 & 0.80 & 0.66 & 0.84 & 0.31 & 0.50 \\ 
Transfusion \citep{zhou2024transfusion} & 8 & 0.67 & - & - & - & - & - & - \\
\midrule
SD3 \citep{esser2024scaling} & 2 & 0.62 & 0.98 &\cellcolor{gray!20} 0.74 & 0.63 & \cellcolor{gray!20}0.67 & 0.34 & \cellcolor{gray!20}0.36 \\
D-DiT (ours) & 2 & 0.65 & 0.97 & \cellcolor{gray!20}0.80 & 0.54 & \cellcolor{gray!20}0.76 & 0.32 & \cellcolor{gray!20}0.50 \\
\bottomrule
\end{tabular}}
\vspace{-2mm}
\caption{Evaluation of text-to-image generation performance on Geneval \citep{ghosh2023geneval}. \textit{params} denote the number of trainable parameters.
}
\end{table}

\begin{table*}
\renewcommand{\arraystretch}{0.98}
  \begin{minipage}[b]{\linewidth}
    \centering
    \resizebox{\linewidth}{!}{
        \begin{tabular}{lcccccccccc}
            \toprule
            \textbf{Model} & \textbf{Params} & \textbf{Text} & \textbf{Image} & \textbf{MS-COCO} & \textbf{VQAv2} & \textbf{VizWiz} & \textbf{OKVQA} & \textbf{MME} & \textbf{GQA} & \textbf{POPE} \\
            & \# trainable & Backbone & Backbone & CIDEr~$\uparrow$ & Acc.~$\uparrow$ & Acc.~$\uparrow$ & Acc.~$\uparrow$ & Acc.~$\uparrow$ & Acc.~$\uparrow$ & Acc.~$\uparrow$\\
            \hline
            InternVL-2.0 \citep{chen2025internvl2}  & 8B & AR & - & - &  - &  \cellcolor{gray!20} 62.9 & \cellcolor{gray!20} 62.9 & \cellcolor{gray!20} 1648.1 &  61.0 & \cellcolor{gray!20} 86.9 \\
            
            LLaVA-Next \citep{liu2024llavanext} & 13B & AR & - & - & \cellcolor{gray!20} 82.8 & 60.5 & - &  1575.0 &  \cellcolor{gray!20} 65.4 & 86.2 \\
            
            BLIP-2 \citep{li2023blip} & 13B & AR & - & - & 65.0 & 19.6 & - & 1293.8 & 41.0 & 85.5 \\
            IDEFICS \citep{laurençon2023idefics} & 9B & AR & - & - & 50.9 & - & - & - & - & - \\
            QWEN-VL \citep{bai2023qwenvl} & 7B & AR & - & - & 78.2 & 38.9 & - & 1487.5 & 57.5 & - \\
            OpenFlamingo \citep{awadalla2023openflamingo} & 9B & AR & - & 65.5 & 43.5 & - & - & - & - & - \\
            Flamingo \citep{alayrac2022flamingo} & 9B & AR & - & \cellcolor{gray!20} 79.4 & 51.8 & 28.8 & 44.7 & - & - & - \\
            \hline
            CM3Leon \citep{yu2023cm3leon}  & 7B & AR & AR & \cellcolor{gray!20}61.6 & 47.6 & \cellcolor{gray!20} 37.6 & 23.8 & - & - & - \\
            Chameleon \citep{team2024chameleon} & 7B & AR & AR & 18.0 & - & - & - & - & - & - \\
            LWM \citep{liu2024world} & 7B & AR & AR & - & 55.8 & 11.6 & - & - & 44.8 & 75.2 \\
            Show-O (256$\times$256) \citep{xie2024show} & 1.3B & AR & Diffusion & - & 64.7 & - & - & 1014.9 &  54.2& 76.2 \\
            Show-O (512$\times$512) \citep{xie2024show} & 1.3B & AR & Diffusion & - &  \cellcolor{gray!20}69.4& - & - & 1097.2 & 58.0 & 80.0 \\
            Transfusion \citep{zhou2024transfusion} & 7B & AR & Diffusion & 29.0 & - & - & - & - & - & - \\
            \hline
            D-DiT (Ours, 256$\times$256) & 2B & \cellcolor{gray!20}Diffusion & \cellcolor{gray!20}Diffusion & - & 59.5 & 19.4 & \cellcolor{gray!20} 28.5 & 897.5 & 55.1 & 79.2 \\
            D-DiT (Ours, 512$\times$512) & 2B & \cellcolor{gray!20}Diffusion & \cellcolor{gray!20}Diffusion & 56.2 & 60.1 & 29.9 & 25.3 & \cellcolor{gray!20}1124.7 & \cellcolor{gray!20}59.2 & \cellcolor{gray!20}84.0 \\
            \bottomrule
        \end{tabular}
    }
    \caption{Comparison of our D-DiT against related work on visual question answering benchmarks. VLMs that focus on text-generation remain superior to unified understanding and generation models, however our models compare favorably with the latter category.}
    \label{table:lm_benchmarks}
  \end{minipage}
  \vspace{-2mm}
\end{table*}

\subsection{Ablation Studies}
As text-to-image diffusion models are trained on a large number of text-image pairs, one may raise the question of whether the representation learned throughout this process can be transferred to multi-modal understanding tasks. To answer this question, we perform an ablation study on the internal representation of a text-to-image diffusion model. We adapt several models into an image captioning model, including SD3, CLIP ViT L/14, and our D-DiT model. Among different internal layers in SD3, we find the feature from the 18th layer tends to perform the best in our preliminary experiments so it is used as the output feature. We add a GPT2 text decoder to the features extracted from SD3 and CLIP, and directly use D-DiT's text output as results. We train all models with a mixture of recaptioned Datacomp, recaptioned OpenImages and captioning data from ShareGPT4V \citep{chen2023sharegpt4v}. Concretely, we evaluate the quality of the captions generated from the models by asking GPT4 to do visual question answering according to the generated captions. The accuracy is listed in Table \ref{tab:diff feature ablation}. 

Similar to the trend observed in \citep{tong2024cambrian}, directly using diffusion features as the prefix of a language decoder yields worse performance compared to language-supervised vision models like CLIP ViT~\citep{radford2021learning}. Unfreezing the parameters of the diffusion backbone slightly improves the performance, but it still cannot match the performance of the CLIP encoder. This suggests that the representation from image diffusion models is not directly transferable to the text embedding space where the decoder-only language model operates on. 
Instead of leveraging a separate language decoder, we use the text branch in the MM-DiT architecture to directly model the conditional text distribution, which notably boosts the performance. This uncovers an intriguing property of MM-DiT models, and potentially other bi-directional Transformers: that these models are good representation learners for estimating the likelihood of multi-modality data distributions. 

We also conduct an ablation study with respect to the number of text diffusion sampling steps and study its influence on VQA accuracy with VQAv2 and captioning quality on the COCO dataset. For VQAv2, which involves short text answers, good accuracy can be achieved with relatively few sampling steps. For the captioning task on MS-COCO, performance improves as the number of sampling steps increases, mirroring the trend observed by \citet{sahoo2024simple}, where additional sampling steps lead to reduced perplexity.

\begin{table}[h]
\centering
\begin{tabular}{cccc} 
\toprule
\multirow{2}{*}{\begin{tabular}[c]{@{}c@{}}\textbf{Vision} \\\textbf{Encoder}\end{tabular}} & \multirow{2}{*}{\begin{tabular}[c]{@{}c@{}}\textbf{Language} \\\textbf{Decoder}\end{tabular}} & \multicolumn{2}{c}{\textbf{VQAv2 (val)}} \\ 
\cmidrule{3-4}
 &  & 0-shot & 32-shot \\
\midrule
SD3 feature (frozen) & GPT 2 & 42.3 & 46.9 \\
SD3 feature (trainable) & GPT 2 & 45.1 & 50.2 \\
CLIP ViT L/14 (frozen) & GPT 2 & 50.6 & 54.8 \\
\midrule
UniDiffuser \citep{bao2023unidiffuser} & GPT 2* & 46.7 & 49.4 \\ 
{D-DiT (ours)} & - & \textbf{55.0} & \textbf{60.3} \\
\bottomrule
\end{tabular}
\vspace{-3mm}
\caption{Comparison between different vision encoder and proposed model. *The GPT 2 decoder of UniDiffuser is finetuned on text reconstruction and kept frozen afterwards.}
\label{tab:diff feature ablation}
\end{table}

\begin{table}[h]
\centering
\setlength{\tabcolsep}{2.8pt}
\resizebox{\linewidth}{!}{
\begin{tabular}{lcccccc}
        \toprule
        \textbf{Task} & $T=4$ & $8$ & $16$ & $32$ & $64$ & $128$ \\
        \hline
        VQAV2 (acc.) & 58.8 & 58.0 & 59.3 & 60.5 & 60.0 & 59.6 \\
        MS-COCO (CIDEr) & 20.2 & 35.3 & 46.5 & 51.3 & 56.2 & 54.5 \\
        \bottomrule
\end{tabular}
}
\vspace{-2mm}
\caption{An ablation study on the effect of sampling steps $T$ on discrete text diffusion performance in terms of COCO Captioning CIDEr score and VQAV2 subset's question-answering accuracy.}
\label{tab:diff sampling steps}
\end{table}
\vspace{-3mm}

\section{Related Works}
\label{sec:related works}

\subsection{Diffusion Models}
Diffusion models \citep{sohl2015deep, ho2020denoising, song2020score} generate data by gradually converting noise into signal via a reverse diffusion process. They are the \textit{de facto} standard for image generation \citep{dhariwal2021diffusion, karras2022elucidating} and likelihood modeling \citep{kingma2021variational, nichol2021improved, song2021maximum,li2024likelihood}. Conditional diffusion models \citep{ho2022classifier} have also been shown to be powerful interfaces bridging text and images, particularly for their ability to generate highly realistic and aesthetic images from textual descriptions \citep{rombach2022ldm, saharia2022imagegen, podell2023sdxl, gao2024lumina, chen2023pixart, ramesh2022dalle2}. Their exceptional performance in the image domain has also inspired numerous extensions to language generation \citep{li2022diffusion,chen2022analog,dieleman2022continuous,lovelace2024latent,gulrajani2024likelihood,lou2023discrete,sahoo2024simple,gat2024discrete,shi2024maskdiff}, and is an attractive alternative as its sampling is not constrained by a specified token generation order and the attention mechanism does not need to be uni-directional.

\subsection{Vision Language Models}
The success of large language models (LLMs) \citep{brown2020language, touvron2023llama} and vision-language pretraining \citep{radford2021learning} has given rise to a series of multi-modal language models. The visual signal is projected to the text embedding space via vision encoders supervised by text labels  \citep{radford2021learning, zhai2023siglip} and then connected to a pretrained language model through further instruction tuning \citep{liu2024visual, zhu2023minigpt, dai2023instructblip, alayrac2022flamingo, tong2024cambrian}. While these models have shown promising capabilities in image understanding and few-shot generalization, their predictive targets are inherently language-centric, limiting their ability to model the image distributions directly.

\subsection{Multimodal Text and Image Generative Models}
Rather than simply connecting visual encoders to language models, recently there has been an active line of inquiry focused on exploring a unified generative model for joint vision and language generation. Inspired by autoregressive language models, many of the unified multi-modal generative models extend the next-token prediction to both image and text tokens \citep{sun2023emu, wang2024emu3, ge2023seed, dong2024dreamllm, team2024chameleon}. More recently, Transfusion \citep{zhou2024transfusion} and Show-O \citep{xie2024show} demonstrate that bi-directional image diffusion can be integrated with autoregressive text prediction in the same framework. 
On the other hand, Versatile Diffusion \citep{xu2023versatile} and Uni-diffuser \citep{bao2023unidiffuser} explore applying a continuous diffusion process to text and image modalities, where text generation is broken into two stages - first, continuous diffusion is used to generate latent embeddings which are then decoded into text by another LLM (e.g. GPT2 \citep{Radford2018ImprovingLU}). While these works hint at the potential of diffusion models as efficient multi-modal models, their text generation capability is restricted to simple tasks like generating short captions from images.
\section*{Conclusion and Discussion}
In this work, we introduced an end-to-end multi-modal diffusion model that bridges the gap between text and image diffusion by enabling both text-to-image (T2I) and image-to-text (I2T) tasks through a unified diffusion model. 
We demonstrated that a bi-directional transformer trained with a joint diffusion target is an effective multi-modal learner capable of competing with the autoregressive models that have long dominated the field. Additionally, the bi-directional attention mechanism is equivariant to the order of input tokens, enabling the prediction of conditional distributions without requiring a specific arrangement of different modalities or special handling of the attention mask.



\clearpage
\newpage
{
    \small
    \bibliographystyle{ieeenat_fullname}
    \bibliography{main}

\begin{thebibliography}{89}
\providecommand{\natexlab}[1]{#1}
\providecommand{\url}[1]{\texttt{#1}}
\expandafter\ifx\csname urlstyle\endcsname\relax
  \providecommand{\doi}[1]{doi: #1}\else
  \providecommand{\doi}{doi: \begingroup \urlstyle{rm}\Url}\fi

\bibitem[Achiam et~al.(2023)Achiam, Adler, Agarwal, Ahmad, Akkaya, Aleman, Almeida, Altenschmidt, Altman, Anadkat, et~al.]{achiam2023gpt}
Josh Achiam, Steven Adler, Sandhini Agarwal, Lama Ahmad, Ilge Akkaya, Florencia~Leoni Aleman, Diogo Almeida, Janko Altenschmidt, Sam Altman, Shyamal Anadkat, et~al.
\newblock Gpt-4 technical report.
\newblock \emph{arXiv preprint arXiv:2303.08774}, 2023.

\bibitem[Alayrac et~al.(2022)Alayrac, Donahue, Luc, Miech, Barr, Hasson, Lenc, Mensch, Millican, Reynolds, et~al.]{alayrac2022flamingo}
Jean-Baptiste Alayrac, Jeff Donahue, Pauline Luc, Antoine Miech, Iain Barr, Yana Hasson, Karel Lenc, Arthur Mensch, Katherine Millican, Malcolm Reynolds, et~al.
\newblock Flamingo: a visual language model for few-shot learning.
\newblock \emph{Advances in neural information processing systems}, 35:\penalty0 23716--23736, 2022.

\bibitem[Anderson(1982)]{anderson1982reverse}
Brian~DO Anderson.
\newblock Reverse-time diffusion equation models.
\newblock \emph{Stochastic Processes and their Applications}, 12\penalty0 (3):\penalty0 313--326, 1982.

\bibitem[Austin et~al.(2021)Austin, Johnson, Ho, Tarlow, and Van Den~Berg]{austin2021structured}
Jacob Austin, Daniel~D Johnson, Jonathan Ho, Daniel Tarlow, and Rianne Van Den~Berg.
\newblock Structured denoising diffusion models in discrete state-spaces.
\newblock \emph{Advances in Neural Information Processing Systems}, 34:\penalty0 17981--17993, 2021.

\bibitem[Awadalla et~al.(2023)Awadalla, Gao, Gardner, Hessel, Hanafy, Zhu, Marathe, Bitton, Gadre, Sagawa, Jitsev, Kornblith, Koh, Ilharco, Wortsman, and Schmidt]{awadalla2023openflamingo}
Anas Awadalla, Irena Gao, Josh Gardner, Jack Hessel, Yusuf Hanafy, Wanrong Zhu, Kalyani Marathe, Yonatan Bitton, Samir Gadre, Shiori Sagawa, Jenia Jitsev, Simon Kornblith, Pang~Wei Koh, Gabriel Ilharco, Mitchell Wortsman, and Ludwig Schmidt.
\newblock Openflamingo: An open-source framework for training large autoregressive vision-language models, 2023.

\bibitem[Bai et~al.(2023)Bai, Bai, Yang, Wang, Tan, Wang, Lin, Zhou, and Zhou]{bai2023qwenvl}
Jinze Bai, Shuai Bai, Shusheng Yang, Shijie Wang, Sinan Tan, Peng Wang, Junyang Lin, Chang Zhou, and Jingren Zhou.
\newblock Qwen-vl: A versatile vision-language model for understanding, localization, text reading, and beyond, 2023.

\bibitem[Bao et~al.(2023)Bao, Nie, Xue, Li, Pu, Wang, Yue, Cao, Su, and Zhu]{bao2023unidiffuser}
Fan Bao, Shen Nie, Kaiwen Xue, Chongxuan Li, Shi Pu, Yaole Wang, Gang Yue, Yue Cao, Hang Su, and Jun Zhu.
\newblock One transformer fits all distributions in multi-modal diffusion at scale.
\newblock In \emph{International Conference on Machine Learning}, pages 1692--1717. PMLR, 2023.

\bibitem[Bigham et~al.(2010)Bigham, Jayant, Ji, Little, Miller, Miller, Miller, Tatarowicz, White, White, et~al.]{bigham2010vizwiz}
Jeffrey~P Bigham, Chandrika Jayant, Hanjie Ji, Greg Little, Andrew Miller, Robert~C Miller, Robin Miller, Aubrey Tatarowicz, Brandyn White, Samual White, et~al.
\newblock Vizwiz: nearly real-time answers to visual questions.
\newblock In \emph{Proceedings of the 23nd annual ACM symposium on User interface software and technology}, pages 333--342, 2010.

\bibitem[Brown(2020)]{brown2020language}
Tom~B Brown.
\newblock Language models are few-shot learners.
\newblock \emph{arXiv preprint arXiv:2005.14165}, 2020.

\bibitem[Chen et~al.(2023{\natexlab{a}})Chen, Yu, Ge, Yao, Xie, Wu, Wang, Kwok, Luo, Lu, and Li]{chen2023pixart}
Junsong Chen, Jincheng Yu, Chongjian Ge, Lewei Yao, Enze Xie, Yue Wu, Zhongdao Wang, James Kwok, Ping Luo, Huchuan Lu, and Zhenguo Li.
\newblock Pixart-$\alpha$: Fast training of diffusion transformer for photorealistic text-to-image synthesis, 2023{\natexlab{a}}.

\bibitem[Chen et~al.(2023{\natexlab{b}})Chen, Li, Dong, Zhang, He, Wang, Zhao, and Lin]{chen2023sharegpt4v}
Lin Chen, Jinsong Li, Xiaoyi Dong, Pan Zhang, Conghui He, Jiaqi Wang, Feng Zhao, and Dahua Lin.
\newblock Sharegpt4v: Improving large multi-modal models with better captions.
\newblock \emph{arXiv preprint arXiv:2311.12793}, 2023{\natexlab{b}}.

\bibitem[Chen et~al.(2022)Chen, Zhang, and Hinton]{chen2022analog}
Ting Chen, Ruixiang Zhang, and Geoffrey Hinton.
\newblock Analog bits: Generating discrete data using diffusion models with self-conditioning.
\newblock \emph{arXiv preprint arXiv:2208.04202}, 2022.

\bibitem[Chen et~al.(2025)Chen, Wang, Cao, Liu, Gao, Cui, Zhu, Ye, Tian, Liu, Gu, Wang, Li, Ren, Chen, Luo, Wang, Jiang, Wang, He, Shi, Zhang, Lv, Wang, Shao, Chu, Tu, He, Wu, Deng, Ge, Chen, Zhang, Wang, Dou, Lu, Zhu, Lu, Lin, Qiao, Dai, and Wang]{chen2025internvl2}
Zhe Chen, Weiyun Wang, Yue Cao, Yangzhou Liu, Zhangwei Gao, Erfei Cui, Jinguo Zhu, Shenglong Ye, Hao Tian, Zhaoyang Liu, Lixin Gu, Xuehui Wang, Qingyun Li, Yimin Ren, Zixuan Chen, Jiapeng Luo, Jiahao Wang, Tan Jiang, Bo Wang, Conghui He, Botian Shi, Xingcheng Zhang, Han Lv, Yi Wang, Wenqi Shao, Pei Chu, Zhongying Tu, Tong He, Zhiyong Wu, Huipeng Deng, Jiaye Ge, Kai Chen, Kaipeng Zhang, Limin Wang, Min Dou, Lewei Lu, Xizhou Zhu, Tong Lu, Dahua Lin, Yu Qiao, Jifeng Dai, and Wenhai Wang.
\newblock Expanding performance boundaries of open-source multimodal models with model, data, and test-time scaling, 2025.

\bibitem[Dai et~al.(2023)Dai, Li, Li, Tiong, Zhao, Wang, Li, Fung, and Hoi]{dai2023instructblip}
Wenliang Dai, Junnan Li, Dongxu Li, Anthony Meng~Huat Tiong, Junqi Zhao, Weisheng Wang, Boyang Li, Pascale Fung, and Steven Hoi.
\newblock Instructblip: Towards general-purpose vision-language models with instruction tuning, 2023.

\bibitem[Dhariwal and Nichol(2021)]{dhariwal2021diffusion}
Prafulla Dhariwal and Alexander Nichol.
\newblock Diffusion models beat gans on image synthesis.
\newblock \emph{Advances in Neural Information Processing Systems}, 34:\penalty0 8780--8794, 2021.

\bibitem[Dieleman et~al.(2022)Dieleman, Sartran, Roshannai, Savinov, Ganin, Richemond, Doucet, Strudel, Dyer, Durkan, et~al.]{dieleman2022continuous}
Sander Dieleman, Laurent Sartran, Arman Roshannai, Nikolay Savinov, Yaroslav Ganin, Pierre~H Richemond, Arnaud Doucet, Robin Strudel, Chris Dyer, Conor Durkan, et~al.
\newblock Continuous diffusion for categorical data.
\newblock \emph{arXiv preprint arXiv:2211.15089}, 2022.

\bibitem[Dong et~al.(2024)Dong, Han, Peng, Qi, Ge, Yang, Zhao, Sun, Zhou, Wei, Kong, Zhang, Ma, and Yi]{dong2024dreamllm}
Runpei Dong, Chunrui Han, Yuang Peng, Zekun Qi, Zheng Ge, Jinrong Yang, Liang Zhao, Jianjian Sun, Hongyu Zhou, Haoran Wei, Xiangwen Kong, Xiangyu Zhang, Kaisheng Ma, and Li Yi.
\newblock Dream{LLM}: Synergistic multimodal comprehension and creation.
\newblock In \emph{The Twelfth International Conference on Learning Representations}, 2024.

\bibitem[Dubey et~al.(2024)Dubey, Jauhri, Pandey, Kadian, Al-Dahle, Letman, Mathur, Schelten, Yang, Fan, et~al.]{dubey2024llama}
Abhimanyu Dubey, Abhinav Jauhri, Abhinav Pandey, Abhishek Kadian, Ahmad Al-Dahle, Aiesha Letman, Akhil Mathur, Alan Schelten, Amy Yang, Angela Fan, et~al.
\newblock The llama 3 herd of models.
\newblock \emph{arXiv preprint arXiv:2407.21783}, 2024.

\bibitem[Esser et~al.(2021)Esser, Rombach, and Ommer]{esser2021taming}
Patrick Esser, Robin Rombach, and Bjorn Ommer.
\newblock Taming transformers for high-resolution image synthesis.
\newblock In \emph{Proceedings of the IEEE/CVF conference on computer vision and pattern recognition}, pages 12873--12883, 2021.

\bibitem[Esser et~al.(2024)Esser, Kulal, Blattmann, Entezari, M{\"u}ller, Saini, Levi, Lorenz, Sauer, Boesel, et~al.]{esser2024scaling}
Patrick Esser, Sumith Kulal, Andreas Blattmann, Rahim Entezari, Jonas M{\"u}ller, Harry Saini, Yam Levi, Dominik Lorenz, Axel Sauer, Frederic Boesel, et~al.
\newblock Scaling rectified flow transformers for high-resolution image synthesis.
\newblock In \emph{Forty-first International Conference on Machine Learning}, 2024.

\bibitem[Fu et~al.(2023)Fu, Chen, Shen, Qin, Zhang, Lin, Yang, Zheng, Li, Sun, et~al.]{fu2023mme}
Chaoyou Fu, Peixian Chen, Yunhang Shen, Yulei Qin, Mengdan Zhang, Xu Lin, Jinrui Yang, Xiawu Zheng, Ke Li, Xing Sun, et~al.
\newblock Mme: A comprehensive evaluation benchmark for multimodal large language models.
\newblock \emph{arXiv preprint arXiv:2306.13394}, 2023.

\bibitem[Gadre et~al.(2023)Gadre, Ilharco, Fang, Hayase, Smyrnis, Nguyen, Marten, Wortsman, Ghosh, Zhang, Orgad, Entezari, Daras, Pratt, Ramanujan, Bitton, Marathe, Mussmann, Vencu, Cherti, Krishna, Koh, Saukh, Ratner, Song, Hajishirzi, Farhadi, Beaumont, Oh, Dimakis, Jitsev, Carmon, Shankar, and Schmidt]{gadre2023datacomp}
Samir~Yitzhak Gadre, Gabriel Ilharco, Alex Fang, Jonathan Hayase, Georgios Smyrnis, Thao Nguyen, Ryan Marten, Mitchell Wortsman, Dhruba Ghosh, Jieyu Zhang, Eyal Orgad, Rahim Entezari, Giannis Daras, Sarah Pratt, Vivek Ramanujan, Yonatan Bitton, Kalyani Marathe, Stephen Mussmann, Richard Vencu, Mehdi Cherti, Ranjay Krishna, Pang~Wei Koh, Olga Saukh, Alexander Ratner, Shuran Song, Hannaneh Hajishirzi, Ali Farhadi, Romain Beaumont, Sewoong Oh, Alex Dimakis, Jenia Jitsev, Yair Carmon, Vaishaal Shankar, and Ludwig Schmidt.
\newblock Datacomp: In search of the next generation of multimodal datasets, 2023.

\bibitem[Gao et~al.(2024)Gao, Zhuo, Lin, Liu, Chen, Du, Xie, Luo, Qiu, Zhang, et~al.]{gao2024lumina}
Peng Gao, Le Zhuo, Ziyi Lin, Chris Liu, Junsong Chen, Ruoyi Du, Enze Xie, Xu Luo, Longtian Qiu, Yuhang Zhang, et~al.
\newblock Lumina-t2x: Transforming text into any modality, resolution, and duration via flow-based large diffusion transformers.
\newblock \emph{arXiv preprint arXiv:2405.05945}, 2024.

\bibitem[Gat et~al.(2024)Gat, Remez, Shaul, Kreuk, Chen, Synnaeve, Adi, and Lipman]{gat2024discrete}
Itai Gat, Tal Remez, Neta Shaul, Felix Kreuk, Ricky~TQ Chen, Gabriel Synnaeve, Yossi Adi, and Yaron Lipman.
\newblock Discrete flow matching.
\newblock \emph{arXiv preprint arXiv:2407.15595}, 2024.

\bibitem[Ge et~al.(2023)Ge, Ge, Zeng, Wang, and Shan]{ge2023seed}
Yuying Ge, Yixiao Ge, Ziyun Zeng, Xintao Wang, and Ying Shan.
\newblock Planting a seed of vision in large language model.
\newblock \emph{arXiv preprint arXiv:2307.08041}, 2023.

\bibitem[Ghosh et~al.(2023)Ghosh, Hajishirzi, and Schmidt]{ghosh2023geneval}
Dhruba Ghosh, Hanna Hajishirzi, and Ludwig Schmidt.
\newblock Geneval: An object-focused framework for evaluating text-to-image alignment, 2023.

\bibitem[Goyal et~al.(2017)Goyal, Khot, Summers{-}Stay, Batra, and Parikh]{balanced_vqa_v2}
Yash Goyal, Tejas Khot, Douglas Summers{-}Stay, Dhruv Batra, and Devi Parikh.
\newblock Making the {V} in {VQA} matter: Elevating the role of image understanding in {V}isual {Q}uestion {A}nswering.
\newblock In \emph{Conference on Computer Vision and Pattern Recognition (CVPR)}, 2017.

\bibitem[Gulrajani and Hashimoto(2024)]{gulrajani2024likelihood}
Ishaan Gulrajani and Tatsunori~B Hashimoto.
\newblock Likelihood-based diffusion language models.
\newblock \emph{Advances in Neural Information Processing Systems}, 36, 2024.

\bibitem[Ho and Salimans(2022)]{ho2022classifier}
Jonathan Ho and Tim Salimans.
\newblock Classifier-free diffusion guidance.
\newblock \emph{arXiv preprint arXiv:2207.12598}, 2022.

\bibitem[Ho et~al.(2020)Ho, Jain, and Abbeel]{ho2020denoising}
Jonathan Ho, Ajay Jain, and Pieter Abbeel.
\newblock Denoising diffusion probabilistic models.
\newblock \emph{Advances in Neural Information Processing Systems}, 33:\penalty0 6840--6851, 2020.

\bibitem[Huang et~al.(2023)Huang, Sun, Xie, Li, and Liu]{huang2023t2i}
Kaiyi Huang, Kaiyue Sun, Enze Xie, Zhenguo Li, and Xihui Liu.
\newblock T2i-compbench: A comprehensive benchmark for open-world compositional text-to-image generation.
\newblock \emph{Advances in Neural Information Processing Systems}, 36:\penalty0 78723--78747, 2023.

\bibitem[Hudson and Manning(2019)]{hudson2019gqa}
Drew~A Hudson and Christopher~D Manning.
\newblock Gqa: A new dataset for real-world visual reasoning and compositional question answering.
\newblock In \emph{Proceedings of the IEEE/CVF conference on computer vision and pattern recognition}, pages 6700--6709, 2019.

\bibitem[Huh et~al.(2024)Huh, Cheung, Wang, and Isola]{huh2024platonic}
Minyoung Huh, Brian Cheung, Tongzhou Wang, and Phillip Isola.
\newblock The platonic representation hypothesis.
\newblock \emph{arXiv preprint arXiv:2405.07987}, 2024.

\bibitem[Karras et~al.(2022)Karras, Aittala, Aila, and Laine]{karras2022elucidating}
Tero Karras, Miika Aittala, Timo Aila, and Samuli Laine.
\newblock Elucidating the design space of diffusion-based generative models.
\newblock \emph{arXiv preprint arXiv:2206.00364}, 2022.

\bibitem[Kingma et~al.(2021)Kingma, Salimans, Poole, and Ho]{kingma2021variational}
Diederik Kingma, Tim Salimans, Ben Poole, and Jonathan Ho.
\newblock Variational diffusion models.
\newblock \emph{Advances in neural information processing systems}, 34:\penalty0 21696--21707, 2021.

\bibitem[Kingma and Welling(2013)]{kingma2013auto}
Diederik~P Kingma and Max Welling.
\newblock Auto-encoding variational bayes.
\newblock \emph{arXiv preprint arXiv:1312.6114}, 2013.

\bibitem[Kuznetsova et~al.(2020)Kuznetsova, Rom, Alldrin, Uijlings, Krasin, Pont-Tuset, Kamali, Popov, Malloci, Kolesnikov, et~al.]{kuznetsova2020openimages}
Alina Kuznetsova, Hassan Rom, Neil Alldrin, Jasper Uijlings, Ivan Krasin, Jordi Pont-Tuset, Shahab Kamali, Stefan Popov, Matteo Malloci, Alexander Kolesnikov, et~al.
\newblock The open images dataset v4: Unified image classification, object detection, and visual relationship detection at scale.
\newblock \emph{International journal of computer vision}, 128\penalty0 (7):\penalty0 1956--1981, 2020.

\bibitem[Laurençon et~al.(2023)Laurençon, Saulnier, Tronchon, Bekman, Singh, Lozhkov, Wang, Karamcheti, Rush, Kiela, Cord, and Sanh]{laurençon2023idefics}
Hugo Laurençon, Lucile Saulnier, Léo Tronchon, Stas Bekman, Amanpreet Singh, Anton Lozhkov, Thomas Wang, Siddharth Karamcheti, Alexander~M. Rush, Douwe Kiela, Matthieu Cord, and Victor Sanh.
\newblock Obelics: An open web-scale filtered dataset of interleaved image-text documents, 2023.

\bibitem[Li et~al.(2024{\natexlab{a}})Li, Kamko, Akhgari, Sabet, Xu, and Doshi]{li2024playground}
Daiqing Li, Aleks Kamko, Ehsan Akhgari, Ali Sabet, Linmiao Xu, and Suhail Doshi.
\newblock Playground v2. 5: Three insights towards enhancing aesthetic quality in text-to-image generation.
\newblock \emph{arXiv preprint arXiv:2402.17245}, 2024{\natexlab{a}}.

\bibitem[Li et~al.(2024{\natexlab{b}})Li, Basri, and Kluger]{li2024likelihood}
Henry Li, Ronen Basri, and Yuval Kluger.
\newblock Likelihood training of cascaded diffusion models via hierarchical volume-preserving maps.
\newblock In \emph{The Twelfth International Conference on Learning Representations}, 2024{\natexlab{b}}.

\bibitem[Li et~al.(2023{\natexlab{a}})Li, Li, Savarese, and Hoi]{li2023blip}
Junnan Li, Dongxu Li, Silvio Savarese, and Steven Hoi.
\newblock Blip-2: Bootstrapping language-image pre-training with frozen image encoders and large language models.
\newblock In \emph{International conference on machine learning}, pages 19730--19742. PMLR, 2023{\natexlab{a}}.

\bibitem[Li et~al.(2024{\natexlab{c}})Li, Tian, Li, Deng, and He]{li2024autoregressive}
Tianhong Li, Yonglong Tian, He Li, Mingyang Deng, and Kaiming He.
\newblock Autoregressive image generation without vector quantization.
\newblock \emph{arXiv preprint arXiv:2406.11838}, 2024{\natexlab{c}}.

\bibitem[Li et~al.(2022{\natexlab{a}})Li, Thickstun, Gulrajani, Liang, and Hashimoto]{li2022diffusion}
Xiang Li, John Thickstun, Ishaan Gulrajani, Percy~S Liang, and Tatsunori~B Hashimoto.
\newblock Diffusion-lm improves controllable text generation.
\newblock \emph{Advances in Neural Information Processing Systems}, 35:\penalty0 4328--4343, 2022{\natexlab{a}}.

\bibitem[Li et~al.(2022{\natexlab{b}})Li, Thickstun, Gulrajani, Liang, and Hashimoto]{li2022diffusionlm}
Xiang Li, John Thickstun, Ishaan Gulrajani, Percy~S Liang, and Tatsunori~B Hashimoto.
\newblock Diffusion-lm improves controllable text generation.
\newblock \emph{Advances in Neural Information Processing Systems}, 35:\penalty0 4328--4343, 2022{\natexlab{b}}.

\bibitem[Li et~al.(2024{\natexlab{d}})Li, Tu, Hui, Wang, Zhao, Xiao, Ren, Mei, Liu, Zheng, et~al.]{li2024recapdatacomp}
Xianhang Li, Haoqin Tu, Mude Hui, Zeyu Wang, Bingchen Zhao, Junfei Xiao, Sucheng Ren, Jieru Mei, Qing Liu, Huangjie Zheng, et~al.
\newblock What if we recaption billions of web images with llama-3?
\newblock \emph{arXiv preprint arXiv:2406.08478}, 2024{\natexlab{d}}.

\bibitem[Li et~al.(2023{\natexlab{b}})Li, Du, Zhou, Wang, Zhao, and Wen]{Li-hallucination-2023}
Yifan Li, Yifan Du, Kun Zhou, Jinpeng Wang, Wayne~Xin Zhao, and Ji-Rong Wen.
\newblock Evaluating object hallucination in large vision-language models.
\newblock In \emph{The 2023 Conference on Empirical Methods in Natural Language Processing}, 2023{\natexlab{b}}.

\bibitem[Lin et~al.(2014)Lin, Maire, Belongie, Hays, Perona, Ramanan, Doll{\'a}r, and Zitnick]{lin2014mscoco}
Tsung-Yi Lin, Michael Maire, Serge Belongie, James Hays, Pietro Perona, Deva Ramanan, Piotr Doll{\'a}r, and C~Lawrence Zitnick.
\newblock Microsoft coco: Common objects in context.
\newblock In \emph{Computer vision--ECCV 2014: 13th European conference, zurich, Switzerland, September 6-12, 2014, proceedings, part v 13}, pages 740--755. Springer, 2014.

\bibitem[Lipman et~al.(2022)Lipman, Chen, Ben-Hamu, Nickel, and Le]{lipman2022flow}
Yaron Lipman, Ricky~TQ Chen, Heli Ben-Hamu, Maximilian Nickel, and Matt Le.
\newblock Flow matching for generative modeling.
\newblock \emph{arXiv preprint arXiv:2210.02747}, 2022.

\bibitem[Liu et~al.(2024{\natexlab{a}})Liu, Li, Li, Li, Zhang, Shen, and Lee]{liu2024llavanext}
Haotian Liu, Chunyuan Li, Yuheng Li, Bo Li, Yuanhan Zhang, Sheng Shen, and Yong~Jae Lee.
\newblock Llava-next: Improved reasoning, ocr, and world knowledge, 2024{\natexlab{a}}.

\bibitem[Liu et~al.(2024{\natexlab{b}})Liu, Li, Wu, and Lee]{liu2024visual}
Haotian Liu, Chunyuan Li, Qingyang Wu, and Yong~Jae Lee.
\newblock Visual instruction tuning.
\newblock \emph{Advances in neural information processing systems}, 36, 2024{\natexlab{b}}.

\bibitem[Liu et~al.(2024{\natexlab{c}})Liu, Yan, Zaharia, and Abbeel]{liu2024lwm}
Hao Liu, Wilson Yan, Matei Zaharia, and Pieter Abbeel.
\newblock World model on million-length video and language with blockwise ringattention, 2024{\natexlab{c}}.

\bibitem[Liu et~al.(2024{\natexlab{d}})Liu, Yan, Zaharia, and Abbeel]{liu2024world}
Hao Liu, Wilson Yan, Matei Zaharia, and Pieter Abbeel.
\newblock World model on million-length video and language with ringattention.
\newblock \emph{arXiv preprint arXiv:2402.08268}, 2024{\natexlab{d}}.

\bibitem[Liu et~al.(2023)Liu, Gong, and qiang liu]{liu2023flow}
Xingchao Liu, Chengyue Gong, and qiang liu.
\newblock Flow straight and fast: Learning to generate and transfer data with rectified flow.
\newblock In \emph{The Eleventh International Conference on Learning Representations}, 2023.

\bibitem[Lou et~al.(2023)Lou, Meng, and Ermon]{lou2023discrete}
Aaron Lou, Chenlin Meng, and Stefano Ermon.
\newblock Discrete diffusion language modeling by estimating the ratios of the data distribution.
\newblock \emph{arXiv preprint arXiv:2310.16834}, 2023.

\bibitem[Lovelace et~al.(2024)Lovelace, Kishore, Wan, Shekhtman, and Weinberger]{lovelace2024latent}
Justin Lovelace, Varsha Kishore, Chao Wan, Eliot Shekhtman, and Kilian~Q Weinberger.
\newblock Latent diffusion for language generation.
\newblock \emph{Advances in Neural Information Processing Systems}, 36, 2024.

\bibitem[Marino et~al.(2019)Marino, Rastegari, Farhadi, and Mottaghi]{marino2019ok}
Kenneth Marino, Mohammad Rastegari, Ali Farhadi, and Roozbeh Mottaghi.
\newblock Ok-vqa: A visual question answering benchmark requiring external knowledge.
\newblock In \emph{Proceedings of the IEEE/cvf conference on computer vision and pattern recognition}, pages 3195--3204, 2019.

\bibitem[Meng et~al.(2022)Meng, Choi, Song, and Ermon]{meng2022concrete}
Chenlin Meng, Kristy Choi, Jiaming Song, and Stefano Ermon.
\newblock Concrete score matching: Generalized score matching for discrete data.
\newblock \emph{Advances in Neural Information Processing Systems}, 35:\penalty0 34532--34545, 2022.

\bibitem[Nichol and Dhariwal(2021)]{nichol2021improved}
Alexander~Quinn Nichol and Prafulla Dhariwal.
\newblock Improved denoising diffusion probabilistic models.
\newblock In \emph{International conference on machine learning}, pages 8162--8171. PMLR, 2021.

\bibitem[Peebles and Xie(2023)]{peebles2023dit}
William Peebles and Saining Xie.
\newblock Scalable diffusion models with transformers.
\newblock In \emph{Proceedings of the IEEE/CVF International Conference on Computer Vision}, pages 4195--4205, 2023.

\bibitem[Podell et~al.(2023)Podell, English, Lacey, Blattmann, Dockhorn, M{\"u}ller, Penna, and Rombach]{podell2023sdxl}
Dustin Podell, Zion English, Kyle Lacey, Andreas Blattmann, Tim Dockhorn, Jonas M{\"u}ller, Joe Penna, and Robin Rombach.
\newblock Sdxl: Improving latent diffusion models for high-resolution image synthesis.
\newblock \emph{arXiv preprint arXiv:2307.01952}, 2023.

\bibitem[Radford and Narasimhan(2018)]{Radford2018ImprovingLU}
Alec Radford and Karthik Narasimhan.
\newblock Improving language understanding by generative pre-training.
\newblock 2018.

\bibitem[Radford et~al.(2019)Radford, Wu, Child, Luan, Amodei, Sutskever, et~al.]{radford2019language}
Alec Radford, Jeffrey Wu, Rewon Child, David Luan, Dario Amodei, Ilya Sutskever, et~al.
\newblock Language models are unsupervised multitask learners.
\newblock \emph{OpenAI blog}, 1\penalty0 (8):\penalty0 9, 2019.

\bibitem[Radford et~al.(2021)Radford, Kim, Hallacy, Ramesh, Goh, Agarwal, Sastry, Askell, Mishkin, Clark, et~al.]{radford2021learning}
Alec Radford, Jong~Wook Kim, Chris Hallacy, Aditya Ramesh, Gabriel Goh, Sandhini Agarwal, Girish Sastry, Amanda Askell, Pamela Mishkin, Jack Clark, et~al.
\newblock Learning transferable visual models from natural language supervision.
\newblock In \emph{International conference on machine learning}, pages 8748--8763. PMLR, 2021.

\bibitem[Raffel et~al.(2023)Raffel, Shazeer, Roberts, Lee, Narang, Matena, Zhou, Li, and Liu]{raffel2023t5}
Colin Raffel, Noam Shazeer, Adam Roberts, Katherine Lee, Sharan Narang, Michael Matena, Yanqi Zhou, Wei Li, and Peter~J. Liu.
\newblock Exploring the limits of transfer learning with a unified text-to-text transformer, 2023.

\bibitem[Ramesh et~al.(2022)Ramesh, Dhariwal, Nichol, Chu, and Chen]{ramesh2022dalle2}
Aditya Ramesh, Prafulla Dhariwal, Alex Nichol, Casey Chu, and Mark Chen.
\newblock Hierarchical text-conditional image generation with clip latents.
\newblock \emph{arXiv preprint arXiv:2204.06125}, 1\penalty0 (2):\penalty0 3, 2022.

\bibitem[Rombach et~al.(2022)Rombach, Blattmann, Lorenz, Esser, and Ommer]{rombach2022ldm}
Robin Rombach, Andreas Blattmann, Dominik Lorenz, Patrick Esser, and Bj{\"o}rn Ommer.
\newblock High-resolution image synthesis with latent diffusion models.
\newblock In \emph{Proceedings of the IEEE/CVF conference on computer vision and pattern recognition}, pages 10684--10695, 2022.

\bibitem[Saharia et~al.(2022)Saharia, Chan, Saxena, Li, Whang, Denton, Ghasemipour, Gontijo~Lopes, Karagol~Ayan, Salimans, et~al.]{saharia2022imagegen}
Chitwan Saharia, William Chan, Saurabh Saxena, Lala Li, Jay Whang, Emily~L Denton, Kamyar Ghasemipour, Raphael Gontijo~Lopes, Burcu Karagol~Ayan, Tim Salimans, et~al.
\newblock Photorealistic text-to-image diffusion models with deep language understanding.
\newblock \emph{Advances in neural information processing systems}, 35:\penalty0 36479--36494, 2022.

\bibitem[Sahoo et~al.(2024)Sahoo, Arriola, Schiff, Gokaslan, Marroquin, Chiu, Rush, and Kuleshov]{sahoo2024simple}
Subham~Sekhar Sahoo, Marianne Arriola, Yair Schiff, Aaron Gokaslan, Edgar Marroquin, Justin~T Chiu, Alexander Rush, and Volodymyr Kuleshov.
\newblock Simple and effective masked diffusion language models.
\newblock \emph{arXiv preprint arXiv:2406.07524}, 2024.

\bibitem[Shi et~al.(2024)Shi, Han, Wang, Doucet, and Titsias]{shi2024maskdiff}
Jiaxin Shi, Kehang Han, Zhe Wang, Arnaud Doucet, and Michalis~K. Titsias.
\newblock Simplified and generalized masked diffusion for discrete data, 2024.

\bibitem[Sohl-Dickstein et~al.(2015)Sohl-Dickstein, Weiss, Maheswaranathan, and Ganguli]{sohl2015deep}
Jascha Sohl-Dickstein, Eric Weiss, Niru Maheswaranathan, and Surya Ganguli.
\newblock Deep unsupervised learning using nonequilibrium thermodynamics.
\newblock In \emph{International Conference on Machine Learning}, pages 2256--2265. PMLR, 2015.

\bibitem[Song et~al.(2020)Song, Sohl-Dickstein, Kingma, Kumar, Ermon, and Poole]{song2020score}
Yang Song, Jascha Sohl-Dickstein, Diederik~P Kingma, Abhishek Kumar, Stefano Ermon, and Ben Poole.
\newblock Score-based generative modeling through stochastic differential equations.
\newblock \emph{arXiv preprint arXiv:2011.13456}, 2020.

\bibitem[Song et~al.(2021)Song, Durkan, Murray, and Ermon]{song2021maximum}
Yang Song, Conor Durkan, Iain Murray, and Stefano Ermon.
\newblock Maximum likelihood training of score-based diffusion models.
\newblock \emph{Advances in Neural Information Processing Systems}, 34:\penalty0 1415--1428, 2021.

\bibitem[Sun et~al.(2023)Sun, Yu, Cui, Zhang, Zhang, Wang, Gao, Liu, Huang, and Wang]{sun2023emu}
Quan Sun, Qiying Yu, Yufeng Cui, Fan Zhang, Xiaosong Zhang, Yueze Wang, Hongcheng Gao, Jingjing Liu, Tiejun Huang, and Xinlong Wang.
\newblock Generative pretraining in multimodality.
\newblock \emph{arXiv preprint arXiv:2307.05222}, 2023.

\bibitem[Tang et~al.(2024)Tang, Yang, Zhu, Zeng, and Bansal]{tang2024codi}
Zineng Tang, Ziyi Yang, Chenguang Zhu, Michael Zeng, and Mohit Bansal.
\newblock Any-to-any generation via composable diffusion.
\newblock \emph{Advances in Neural Information Processing Systems}, 36, 2024.

\bibitem[Team(2024)]{team2024chameleon}
Chameleon Team.
\newblock Chameleon: Mixed-modal early-fusion foundation models.
\newblock \emph{arXiv preprint arXiv:2405.09818}, 2024.

\bibitem[Team et~al.(2023)Team, Anil, Borgeaud, Alayrac, Yu, Soricut, Schalkwyk, Dai, Hauth, Millican, et~al.]{team2023gemini}
Gemini Team, Rohan Anil, Sebastian Borgeaud, Jean-Baptiste Alayrac, Jiahui Yu, Radu Soricut, Johan Schalkwyk, Andrew~M Dai, Anja Hauth, Katie Millican, et~al.
\newblock Gemini: a family of highly capable multimodal models.
\newblock \emph{arXiv preprint arXiv:2312.11805}, 2023.

\bibitem[Tian et~al.(2024)Tian, Jiang, Yuan, Peng, and Wang]{tian2024visual}
Keyu Tian, Yi Jiang, Zehuan Yuan, Bingyue Peng, and Liwei Wang.
\newblock Visual autoregressive modeling: Scalable image generation via next-scale prediction.
\newblock \emph{arXiv preprint arXiv:2404.02905}, 2024.

\bibitem[Tong et~al.(2024)Tong, Brown, Wu, Woo, Middepogu, Akula, Yang, Yang, Iyer, Pan, et~al.]{tong2024cambrian}
Shengbang Tong, Ellis Brown, Penghao Wu, Sanghyun Woo, Manoj Middepogu, Sai~Charitha Akula, Jihan Yang, Shusheng Yang, Adithya Iyer, Xichen Pan, et~al.
\newblock Cambrian-1: A fully open, vision-centric exploration of multimodal llms.
\newblock \emph{arXiv preprint arXiv:2406.16860}, 2024.

\bibitem[Touvron et~al.(2023)Touvron, Lavril, Izacard, Martinet, Lachaux, Lacroix, Rozi{\`e}re, Goyal, Hambro, Azhar, et~al.]{touvron2023llama}
Hugo Touvron, Thibaut Lavril, Gautier Izacard, Xavier Martinet, Marie-Anne Lachaux, Timoth{\'e}e Lacroix, Baptiste Rozi{\`e}re, Naman Goyal, Eric Hambro, Faisal Azhar, et~al.
\newblock Llama: Open and efficient foundation language models.
\newblock \emph{arXiv preprint arXiv:2302.13971}, 2023.

\bibitem[Vincent(2011)]{vincent2011connection}
Pascal Vincent.
\newblock A connection between score matching and denoising autoencoders.
\newblock \emph{Neural computation}, 23\penalty0 (7):\penalty0 1661--1674, 2011.

\bibitem[Wang et~al.(2024)Wang, Zhang, Luo, Sun, Cui, Wang, Zhang, Wang, Li, Yu, et~al.]{wang2024emu3}
Xinlong Wang, Xiaosong Zhang, Zhengxiong Luo, Quan Sun, Yufeng Cui, Jinsheng Wang, Fan Zhang, Yueze Wang, Zhen Li, Qiying Yu, et~al.
\newblock Emu3: Next-token prediction is all you need.
\newblock \emph{arXiv preprint arXiv:2409.18869}, 2024.

\bibitem[Wu et~al.(2024)Wu, Zhang, Chen, Tang, Li, Fang, Zhu, Xie, Yin, Yi, et~al.]{wu2024vila}
Yecheng Wu, Zhuoyang Zhang, Junyu Chen, Haotian Tang, Dacheng Li, Yunhao Fang, Ligeng Zhu, Enze Xie, Hongxu Yin, Li Yi, et~al.
\newblock Vila-u: a unified foundation model integrating visual understanding and generation.
\newblock \emph{arXiv preprint arXiv:2409.04429}, 2024.

\bibitem[Xie et~al.(2024)Xie, Mao, Bai, Zhang, Wang, Lin, Gu, Chen, Yang, and Shou]{xie2024show}
Jinheng Xie, Weijia Mao, Zechen Bai, David~Junhao Zhang, Weihao Wang, Kevin~Qinghong Lin, Yuchao Gu, Zhijie Chen, Zhenheng Yang, and Mike~Zheng Shou.
\newblock Show-o: One single transformer to unify multimodal understanding and generation.
\newblock \emph{arXiv preprint arXiv:2408.12528}, 2024.

\bibitem[Xu et~al.(2023)Xu, Wang, Zhang, Wang, and Shi]{xu2023versatile}
Xingqian Xu, Zhangyang Wang, Gong Zhang, Kai Wang, and Humphrey Shi.
\newblock Versatile diffusion: Text, images and variations all in one diffusion model.
\newblock In \emph{Proceedings of the IEEE/CVF International Conference on Computer Vision}, pages 7754--7765, 2023.

\bibitem[Yu et~al.(2023{\natexlab{a}})Yu, Lezama, Gundavarapu, Versari, Sohn, Minnen, Cheng, Birodkar, Gupta, Gu, et~al.]{yu2023language}
Lijun Yu, Jos{\'e} Lezama, Nitesh~B Gundavarapu, Luca Versari, Kihyuk Sohn, David Minnen, Yong Cheng, Vighnesh Birodkar, Agrim Gupta, Xiuye Gu, et~al.
\newblock Language model beats diffusion--tokenizer is key to visual generation.
\newblock \emph{arXiv preprint arXiv:2310.05737}, 2023{\natexlab{a}}.

\bibitem[Yu et~al.(2023{\natexlab{b}})Yu, Shi, Pasunuru, Muller, Golovneva, Wang, Babu, Tang, Karrer, Sheynin, Ross, Polyak, Howes, Sharma, Xu, Tamoyan, Ashual, Singer, Li, Zhang, James, Ghosh, Taigman, Fazel-Zarandi, Celikyilmaz, Zettlemoyer, and Aghajanyan]{yu2023cm3leon}
Lili Yu, Bowen Shi, Ramakanth Pasunuru, Benjamin Muller, Olga Golovneva, Tianlu Wang, Arun Babu, Binh Tang, Brian Karrer, Shelly Sheynin, Candace Ross, Adam Polyak, Russell Howes, Vasu Sharma, Puxin Xu, Hovhannes Tamoyan, Oron Ashual, Uriel Singer, Shang-Wen Li, Susan Zhang, Richard James, Gargi Ghosh, Yaniv Taigman, Maryam Fazel-Zarandi, Asli Celikyilmaz, Luke Zettlemoyer, and Armen Aghajanyan.
\newblock Scaling autoregressive multi-modal models: Pretraining and instruction tuning, 2023{\natexlab{b}}.

\bibitem[Zhai et~al.(2023)Zhai, Mustafa, Kolesnikov, and Beyer]{zhai2023siglip}
Xiaohua Zhai, Basil Mustafa, Alexander Kolesnikov, and Lucas Beyer.
\newblock Sigmoid loss for language image pre-training, 2023.

\bibitem[Zhou et~al.(2024)Zhou, Yu, Babu, Tirumala, Yasunaga, Shamis, Kahn, Ma, Zettlemoyer, and Levy]{zhou2024transfusion}
Chunting Zhou, Lili Yu, Arun Babu, Kushal Tirumala, Michihiro Yasunaga, Leonid Shamis, Jacob Kahn, Xuezhe Ma, Luke Zettlemoyer, and Omer Levy.
\newblock Transfusion: Predict the next token and diffuse images with one multi-modal model.
\newblock \emph{arXiv preprint arXiv:2408.11039}, 2024.

\bibitem[Zhu et~al.(2023)Zhu, Chen, Shen, Li, and Elhoseiny]{zhu2023minigpt}
Deyao Zhu, Jun Chen, Xiaoqian Shen, Xiang Li, and Mohamed Elhoseiny.
\newblock Minigpt-4: Enhancing vision-language understanding with advanced large language models.
\newblock \emph{arXiv preprint arXiv:2304.10592}, 2023.

\end{thebibliography}
}
\clearpage
\setcounter{page}{1}
\maketitlesupplementary


\section{Training Details}
\begin{table}[ht]
\centering
\resizebox{\linewidth}{!}{

\begin{tabular}{ccccc} 
\toprule
\multirow{2}{*}{Hyperparam.} & \multirow{2}{*}{Dual pretrain} & \multicolumn{2}{c}{Continued pretrain} & \multirow{2}{*}{Instruct. tuning} \\ 
\cmidrule{3-4}
 &  & Mask emb. & High res. &  \\ 
\midrule
Gradient steps & 60k & 200k & 80k & 50k \\
Batch size & 512 & 512 & 768 & 512 \\
LR & 5e-5 & 3e-5 & 3e-5 & 3e-5 \\
Scheduler & \multicolumn{4}{c}{Constant LR with warmup} \\
Warmup iters & 5000 & 1000 & 1000 & 1000 \\
Weight decay & \multicolumn{4}{c}{1e-2} \\
Text loss weight & \multicolumn{3}{c}{0.2} & 1.0 \\
\bottomrule
\end{tabular}}
\caption{Training hyperparameters for D-DiT. Text loss weight denotes the $\lambda$ in \eqref{eq:joint diffusion loss}.}
\label{tab:training hyperparam}
\end{table}

We provide the detailed hyperparameter setting for different training stages in the Table \ref{tab:training hyperparam}. During all the training stages, we use AdamW optimizer with default hyperparameters ($\beta_1=0.9, \beta_2=0.999$). Mixed precision training (bf16) and fully-sharded data parallel (with gradient and optimizer state sharded) are used for model training.

\section{Further Results}
\begin{table}[h]
\centering
\begin{tabular}{cccc} 
\toprule
Model & Backbone & Params. (B) & FID $\downarrow$ \\ 
\midrule
SD-XL \citep{podell2023sdxl} & Diff. & 0.9 & 9.55 \\
PixArt-$\alpha$ \citep{chen2023pixart} & Diff. & 0.6 & 6.14 \\
Playground v2.5 & Diff. & - & 4.48 \\
Show-O \citep{xie2024show} & Discrete Diff. & 1.3 & 15.18 \\
LWM \citep{liu2024lwm} & AR & 7 & 17.77 \\
VILA-U \citep{wu2024vila} & AR & 7 & 7.69 \\ 
\midrule
SD3 \citep{esser2024scaling} & Diff. & 2 & 16.45 \\
D-DiT & Diff. & 2 & 15.16 \\
\bottomrule
\end{tabular}
\caption{Comparison with other models
on MJHQ-30K evaluation benchmark at $512\times512$ resolution.}
\label{tab:mjhq fid}
\end{table}

\begin{table}[h]
\centering
\fontsize{9.5pt}{9.5pt}\selectfont
\begin{tabular}{cccccc} 
\toprule
\setlength{\tabcolsep}{2.0pt}
\multirow{2}{*}{Model} & \multicolumn{2}{c}{COCO-30k} & \multicolumn{3}{c}{T2I CompBench} \\ 
\cmidrule(lr){2-3}\cmidrule(lr){4-6}
 & FID $\downarrow$ & CLIP $\uparrow$ & Color $\uparrow$ & Shape $\uparrow$ & Texture $\uparrow$ \\ 
\midrule
SD3 & 10.2 & 30.9 & 0.7993 & 0.5816 & 0.7389 \\
D-DiT & 9.4 & 31.2 & 0.8001 & 0.5703  & 0.6856 \\
\bottomrule
\end{tabular}
\caption{Further image generation comparisons against original SD3 on MS-COCO dataset \citep{lin2014mscoco} and T2I CompBench \citep{huang2023t2i}.}
\label{table:sd3_comparison}
\end{table}

\paragraph{Image generation} We evaluate the aesthetic quality of generated images from our proposed D-DiT against those of the original SD3 model and a selection of existing text-to-image (T2I) and multi-modal works. We measure Frechet Inception Distance (FID) with respect to a collection highly aesthetic generated images, known as the MJHQ-30K benchmark proposed by \citep{li2024playground}. As shown in Table \ref{tab:mjhq fid}, we observe an improvement in FID after joint diffusion training, and favorable comparison against multi-modal models of similar size. We also provide further comparisons on MS-COCO 30k and T2I CompBench in Table \ref{table:sd3_comparison}. The FID and CLIP score slightly improve compared to the original SD3 model. On T2I CompBench, we find that after dual diffusion fine tuning the model performs worse in texture. We hypothesize that the major reason is the texture quality of our training dataset is worse than the dataset used for training SD3.

\begin{figure*}[t]
    \centering
    \includegraphics[width=\linewidth]{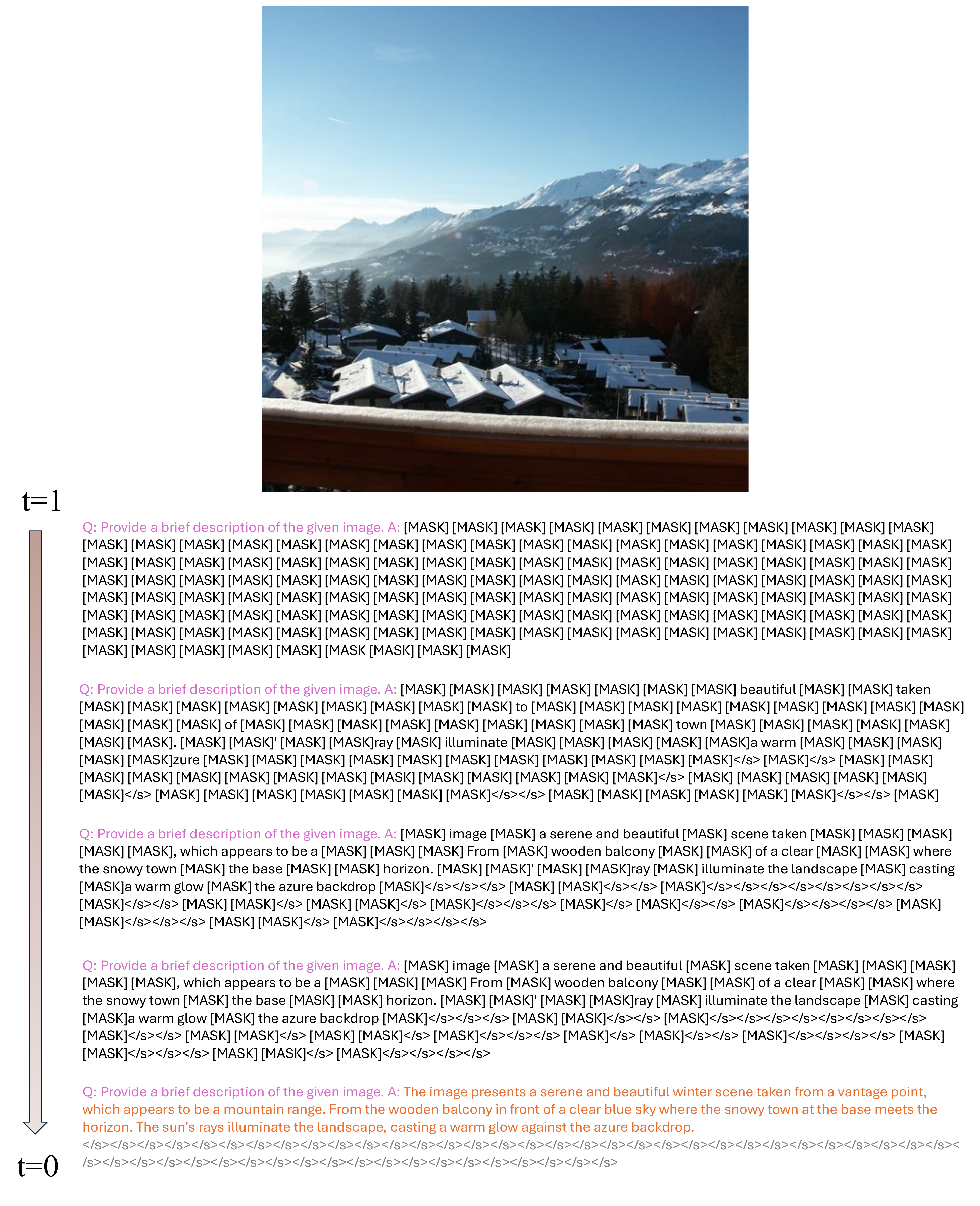}
    \vspace{-1mm}
    \caption{Illustrative example of visual question answering with mask diffusion.}
    \label{fig:supp_text_flow}
    \vspace{-3mm}
\end{figure*}

\paragraph{Text generation process} We provide an illustrative example of masked diffusion in Figure \ref{fig:supp_text_flow} for the visual question answering task, where the token generation process is visualized over diffusion time. Over the course of sampling, the answer tokens are gradually denoised from the masked state via absorbing state reverse diffusion. The question tokens are always left unmasked throughout the entire process.

\begin{table}[h]
\centering
\fontsize{9pt}{9pt}\selectfont
\setlength{\tabcolsep}{2.0pt}
\begin{tabular}{ccccccc} 
\toprule
\multirow{2}{*}{Model} & \multirow{2}{*}{\begin{tabular}[c]{@{}c@{}}\# \\trainable\end{tabular}} & \multirow{2}{*}{\begin{tabular}[c]{@{}c@{}}Text \\encoder\end{tabular}} & \multirow{2}{*}{Geneval} & \multirow{2}{*}{\begin{tabular}[c]{@{}c@{}}COCO \\FID\end{tabular}} & \multicolumn{2}{c}{VQAv2(val)} \\ 
\cmidrule(lr){6-7}
 &  &  &  &  & 0-shot & 32-shot \\ 
\midrule
End-to-End & 1.1B & - & 0.39 & 18.1 & 54.3 & 58.7 \\
From SD3 & 2B & T5-XXL & 0.65 & 9.4 & 55.0 & 60.3 \\
\bottomrule
\end{tabular}
\caption{Comparison of different D-DiT variants.  \textit{End-to-End} variant is trained from scratch and uses GPT2's text tokenizer. \textit{From SD3} variant is  initialized from SD3 pretrained checkpoint and uses T5 encoder. The end-to-end model is first trained on OpenWebText for 350B tokens, then trained on DataComp-recap1B for an epoch (400k steps) and a filtered subset for 100k steps. \label{tab:end2end_ablation}}
\end{table}

\paragraph{Training from scratch and removing T5 encoder} To study the influence of text-to-image pretraining, we conduct a study by comparing a D-DiT model that is trained from scratch. We found that initializing from pretrained text-to-image model and use a pretrained text encoder can greatly aid model learning of text-to-image tasks. Meanwhile, image captioning on VQA also mildly improves (Table \ref{tab:end2end_ablation}).

\paragraph{Image generation's influence on SFT} To analyze the influence of dual diffusion loss on image understanding, we conduct supervised finetune on LLaVA 1.5 dataset with varying amount of image generation data, including a training that only has understanding loss (no generation data). We observe that the image generation loss and corresponding data amount
does not have significant influence on model's understanding performance (Table \ref{table:data_ablation}).

\begin{table}[h]
\centering
\fontsize{9.0pt}{9.0pt}\selectfont
\begin{tabular}{cccccccc} 
\toprule
\multirow{2}{*}{Und.} & \multirow{2}{*}{Gen.} & \multicolumn{3}{c}{VQAv2 (val)} & \multicolumn{3}{c}{POPE} \\ 
\cmidrule(r){3-5}\cmidrule(r){6-8}
 &  & 10k & 30k & 50k & 10k & 30k & 50k \\ 
\midrule
0.665M & 0 & 52.8 & 55.9 & 58.3 & 79.6 & 80.9 & 81.8 \\
0.665M & 7M & 53.4 & 55.8 & 58.1 & 79.8 & 81.2 & 82.4 \\
0.665M & 20M & 53.6 & 55.8 & 58.3 & 81.0  & 81.1 & 82.5 \\
\bottomrule
\end{tabular}
\vspace{-.14in}
\caption{Understanding performance (accuracy) under different data settings and training steps during supervised finetune. Batch size is set to 128 for this experiment.}
\label{table:data_ablation}
\end{table}

\begin{figure*}[t]
    \centering
    \includegraphics[width=0.97\linewidth]{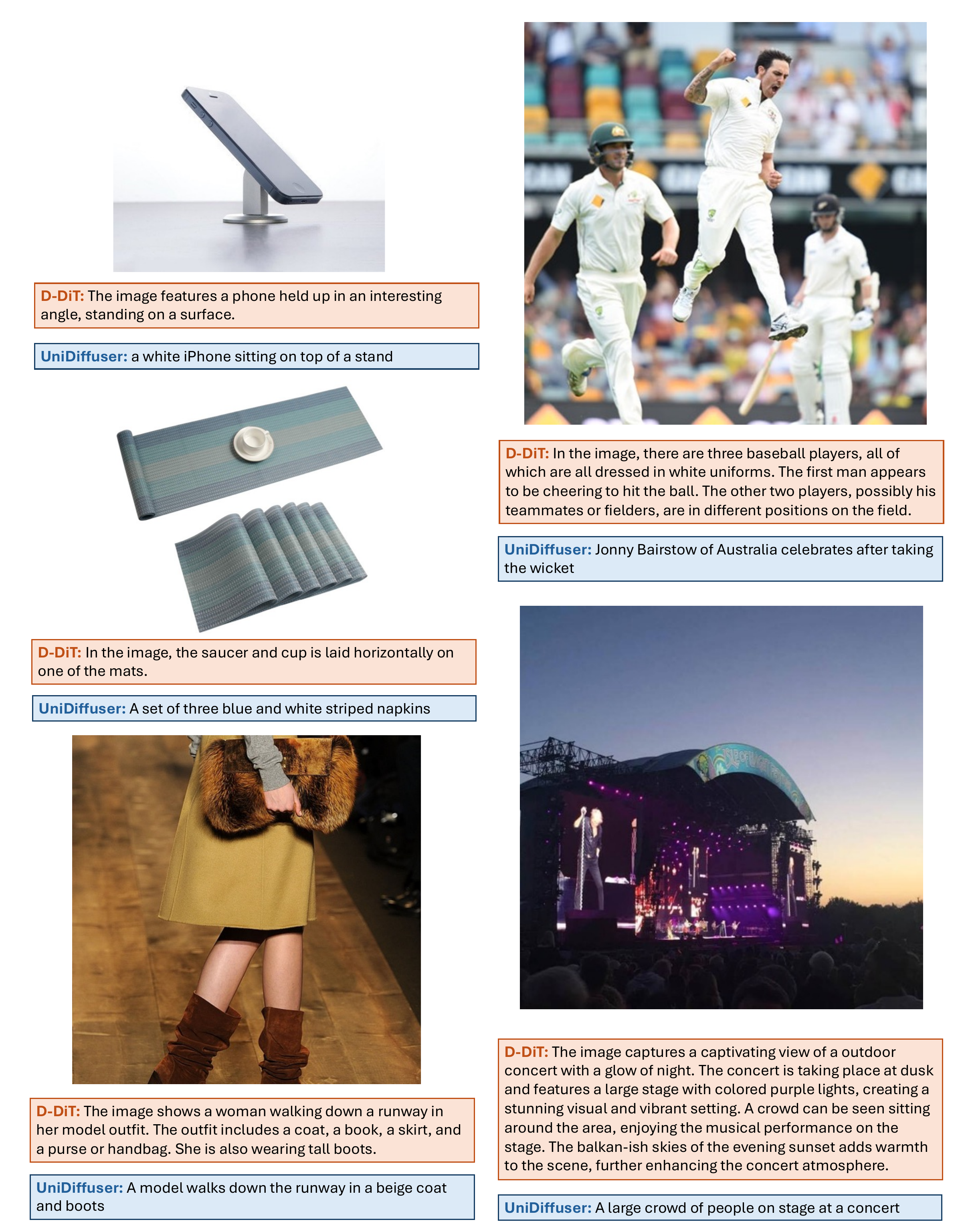}
    \vspace{-1mm}
    \caption{Comparison of captions generated by D-DiT and UniDiffuser\citep{bao2023unidiffuser}. The prompt to D-DiT is "Provide a brief description of the given image."}
    \label{fig:supp_i2t}
    \vspace{-3mm}
\end{figure*}

\begin{figure*}[t]
    \centering
    \includegraphics[width=\linewidth]{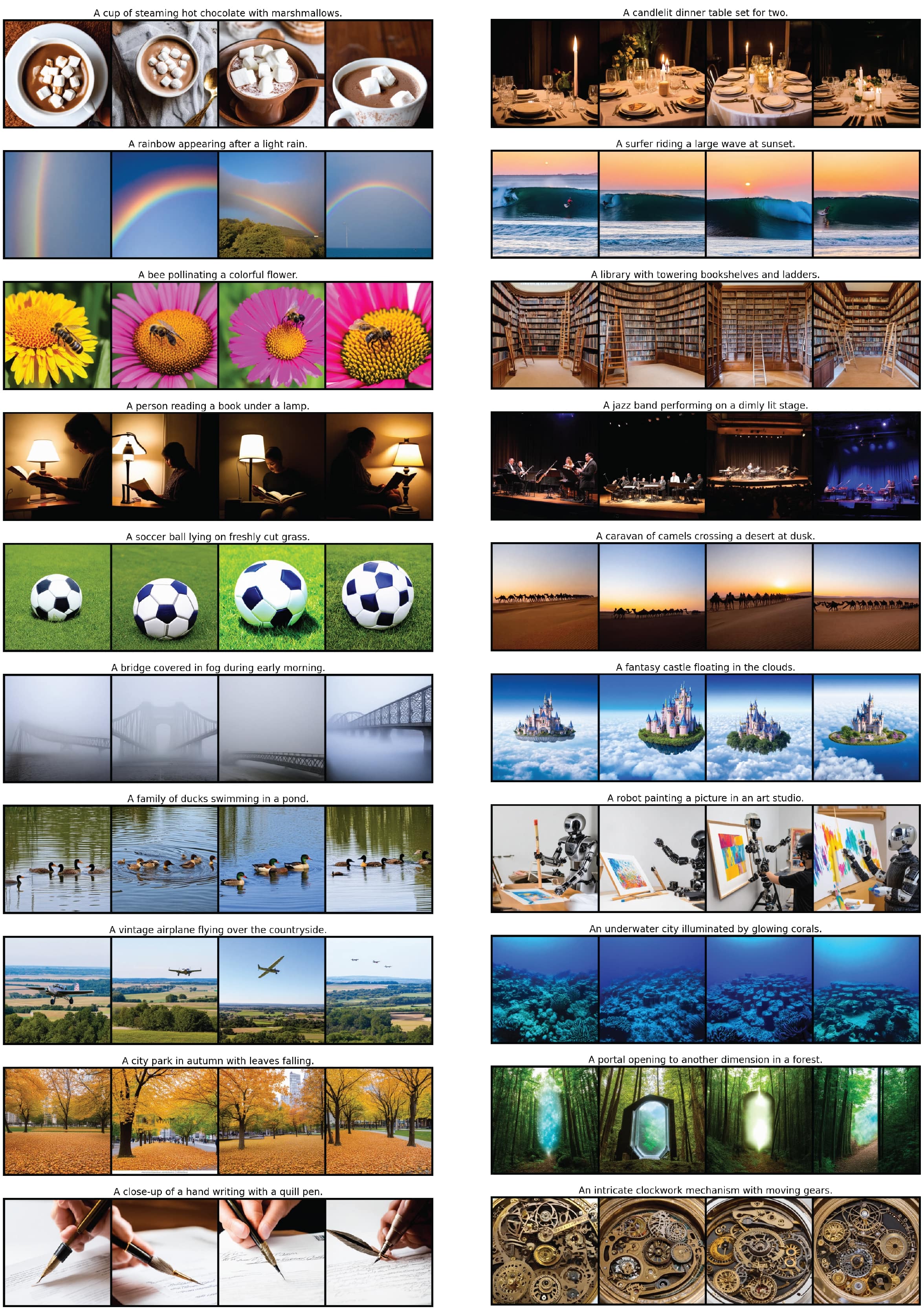}
        \vspace{-7mm}
    \caption{Additional text-to-image samples generated from the model.}
    \label{fig:supp_t2i}
    \vspace{-3mm}
\end{figure*}

\begin{figure*}[t]
    \centering
    \includegraphics[width=\linewidth]{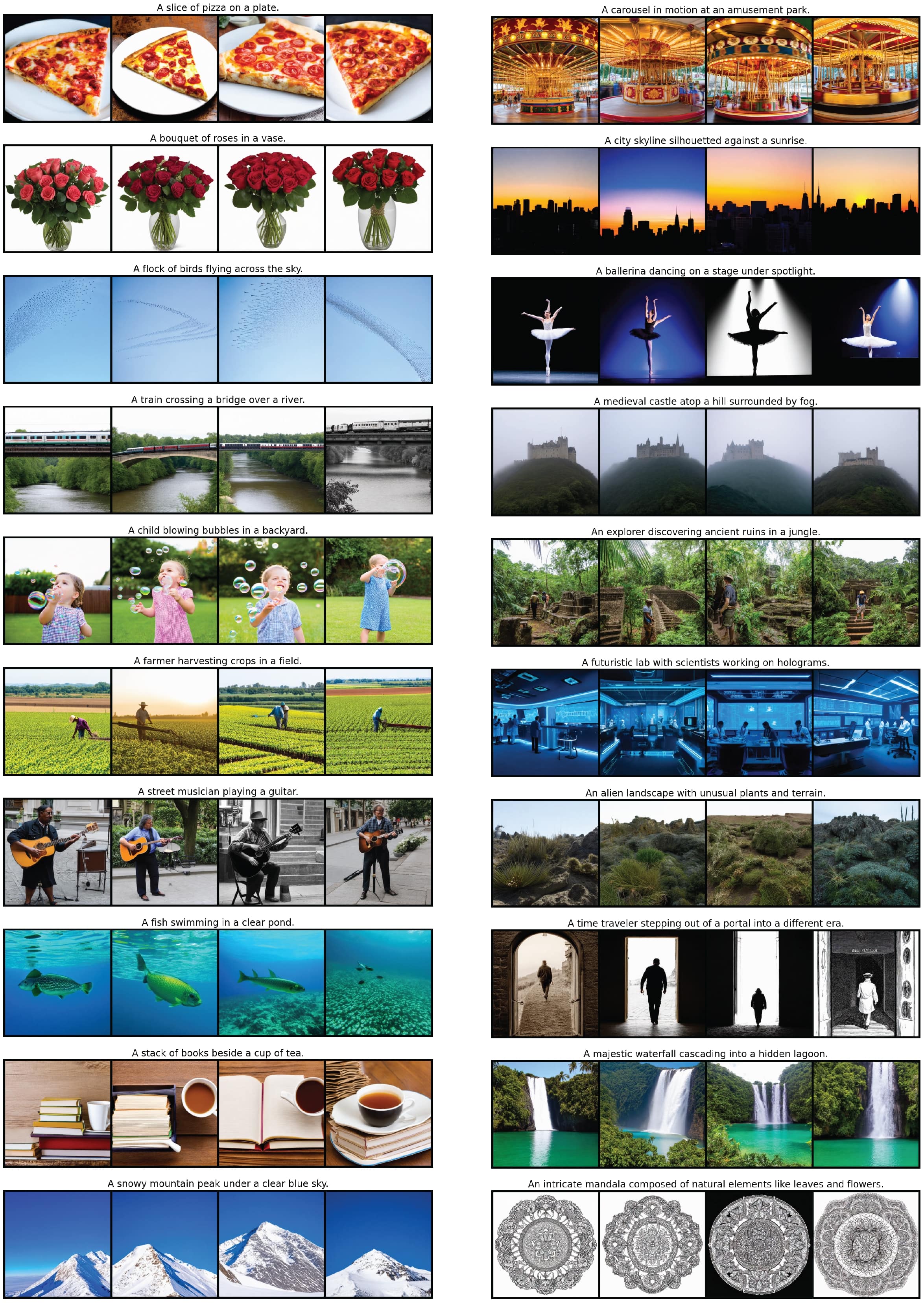}
        \vspace{-7mm}
    \caption{Additional text-to-image samples generated from the model.}
    \label{fig:supp_t2i_2}
    \vspace{-3mm}
\end{figure*}

\begin{figure*}[t]
    \centering
    \includegraphics[width=\linewidth]{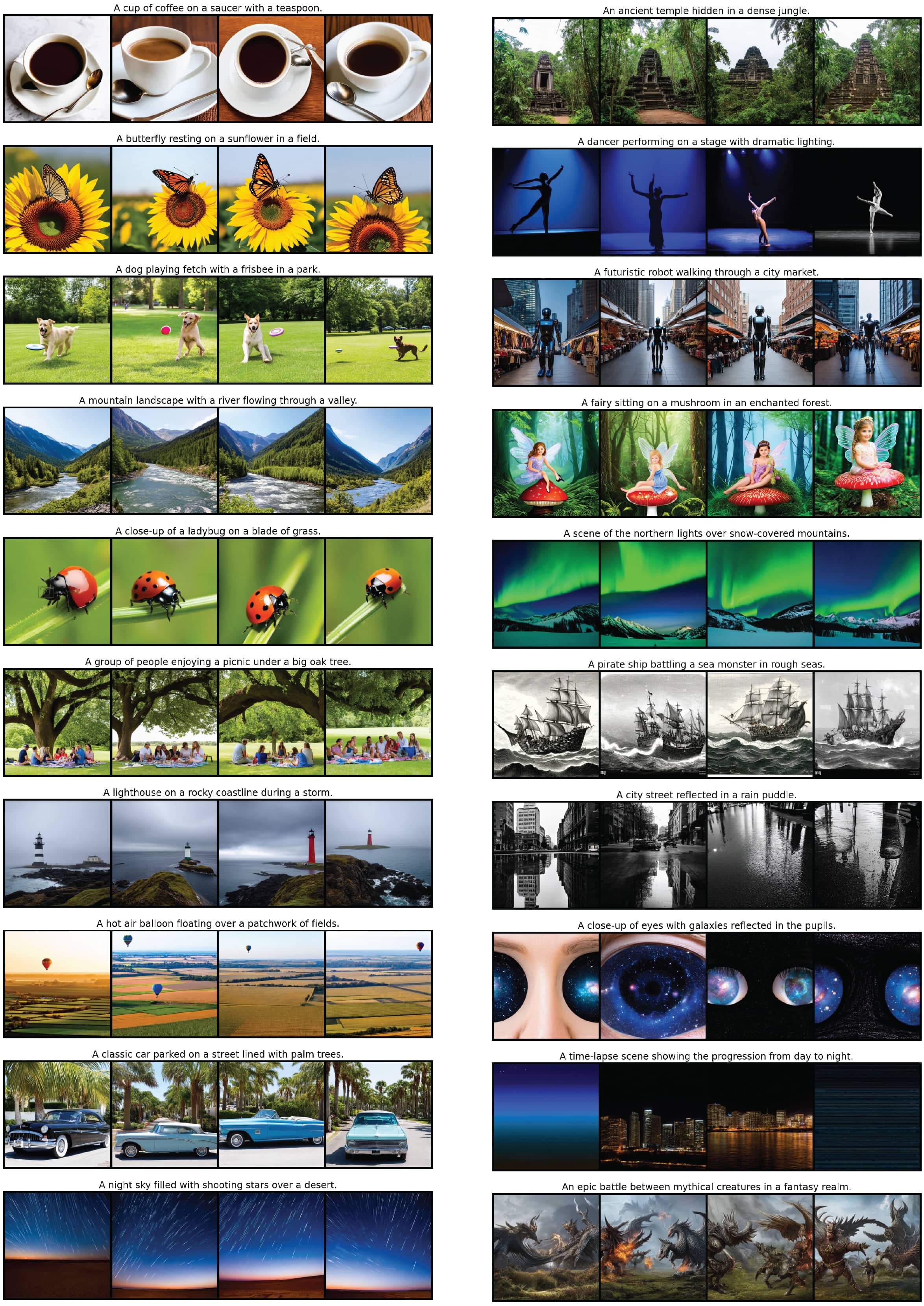}
        \vspace{-7mm}
    \caption{Additional text-to-image samples generated from the model.}
    \label{fig:supp_t2i_3}
    \vspace{-3mm}
\end{figure*}

\begin{figure*}[t]
    \centering
    \includegraphics[width=\linewidth]{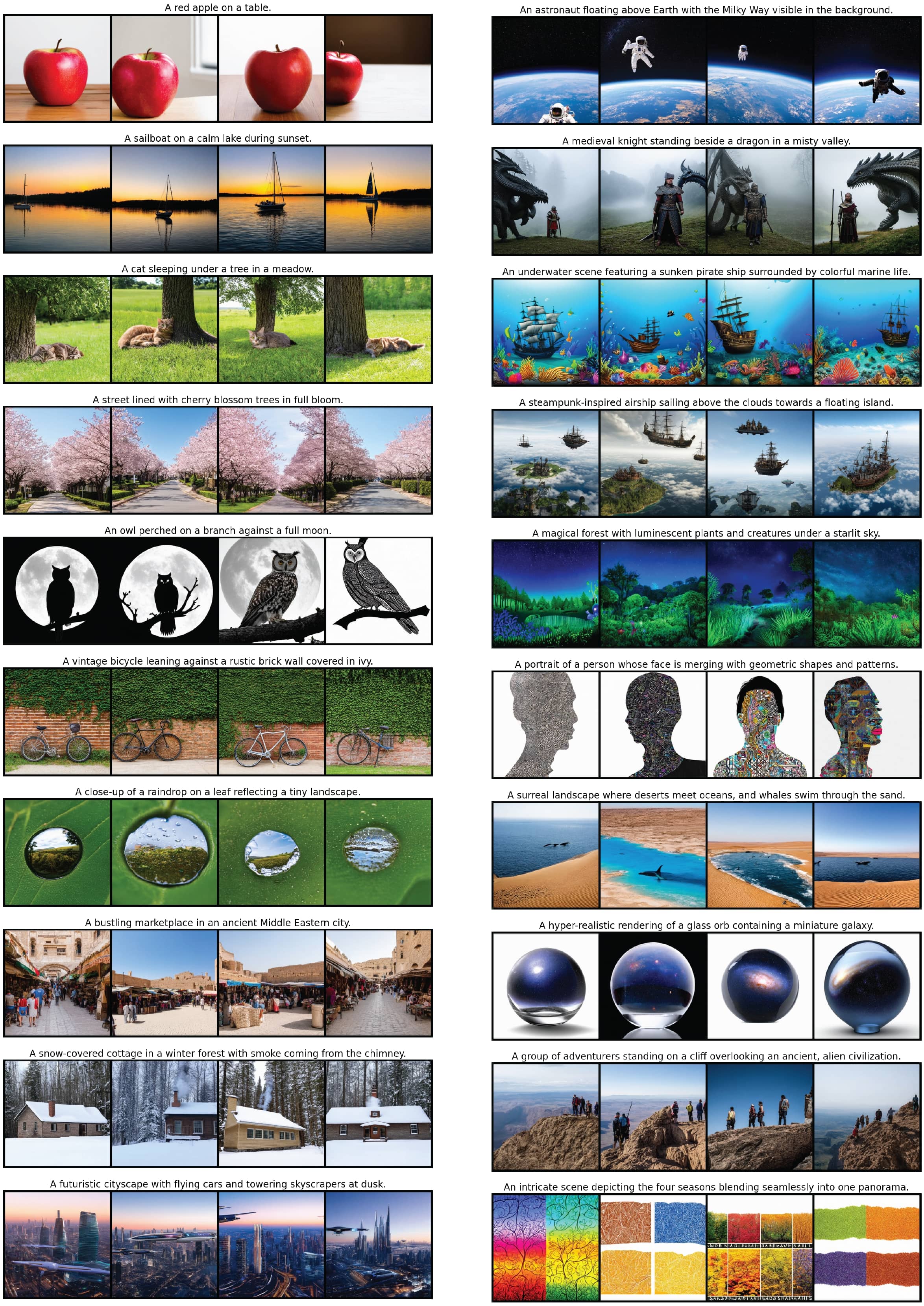}
        \vspace{-7mm}
    \caption{Additional text-to-image samples generated from the model.}
    \label{fig:supp_t2i_4}
    \vspace{-3mm}
\end{figure*}

\paragraph{Comaprison against previous multi-modal diffusion model} We also include a qualitative comparison in captioning performance compared to UniDiffuser \citep{bao2023unidiffuser}, another diffusion-based multi-modal model, in Figure \ref{fig:supp_i2t}, where we demonstrate an improvement in the ability to capture fine-grained details of the image in a longer caption format. Finally, we provide further uncurated text-to-image (T2I) generation results in Figures \ref{fig:supp_t2i}, \ref{fig:supp_t2i_2}, \ref{fig:supp_t2i_3}, and \ref{fig:supp_t2i_4}. Overall, these results further demonstrate the multi-faceted performance of our proposed dual-branch diffusion-based multi-modal model.

\begin{figure*}[t]
\centering
    \begin{subfigure}[B]{0.49\linewidth}
        \centering
\includegraphics[width=0.80\linewidth]{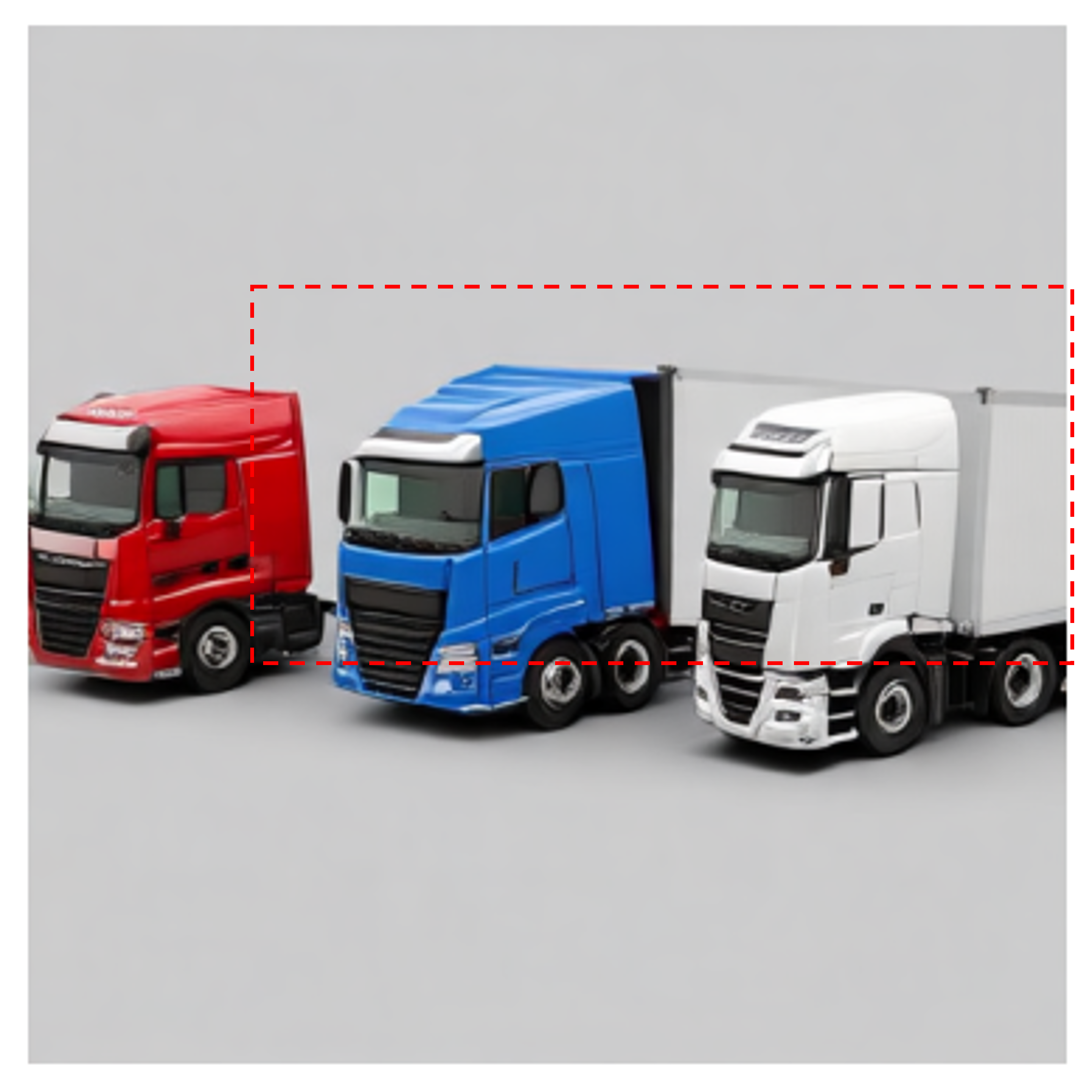}
    \caption{ T2I Prompt: \textit{Three trucks parking in parallel: one red, one blue, and one white. Red truck has load and the rest don't have. }   \label{fig:t2i failure}}
    \end{subfigure}
    \begin{subfigure}[B]{0.5\linewidth}
        \centering
    \includegraphics[width=\linewidth]{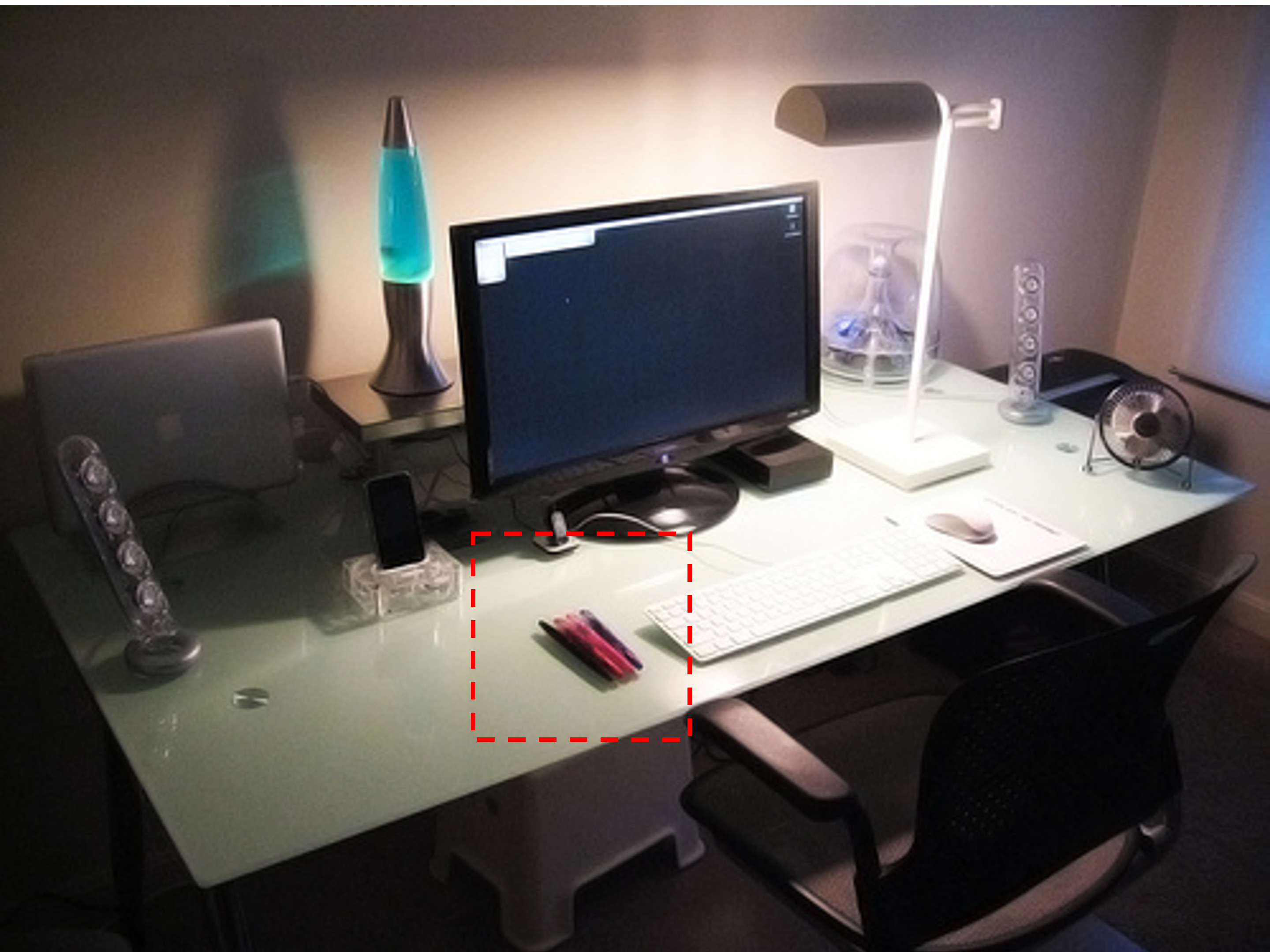}
    \vspace{+0.5mm}
    \caption{ I2T Prompt:\textit{Q: How many pens are there on the desk and what are their colors?} A: There are three pens on the desk, and they are red and blue.    \label{fig:i2t failure}}
    \end{subfigure}
    \caption{Examples of failed text-to-image and image-to-text generation.}
\end{figure*}

\paragraph{Limitations} 

As shown in Figure~\ref{fig:i2t failure}:
in T2I, we find that D-DiT can struggle to generate scenes with relatively complex instructions. In I2T, D-DiT can fail to identify the full details of smaller objects. We also observe model's performance performance deteriorates with longer prompts, primarily due to the bias towards short prompts in the LLaVA finetuning dataset.

In summary, while discrete diffusion offers the advantage of being agnostic to sequential order and is compatible with bi-directional Transformers, its current implementation requires the sequence length to be preset before sampling. A promising future direction would be to extend the sampling scheme to allow for more flexibility, enabling dynamic sequence lengths during the sampling process. 
In addition, while we show that our proposed dual diffusion model can perform instruction tuning, its instruction-following capabilities still marginally lag behind those of state-of-the-art autoregressive models.

\end{document}